%% file: main.tex
\documentclass[10pt,twocolumn,letterpaper]{article}

\usepackage[pagenumbers]{cvpr} 
\usepackage{newfloat}
\usepackage{listings}
\usepackage{graphicx}
\usepackage{pifont} 
\usepackage{comment}
\usepackage{color}
\usepackage{dsfont}
\usepackage{enumitem}

\usepackage{multicol}
\usepackage{times}
\usepackage{epsfig}
\usepackage{graphicx}
\usepackage{subcaption}
\usepackage{booktabs}
\usepackage{array}
\usepackage{caption}
\usepackage{algorithm}
\usepackage{algorithmic}
\usepackage{amsmath, amsthm, amssymb}
\usepackage{threeparttable}
\usepackage[accsupp]{axessibility}
\usepackage{wasysym}
\usepackage{colortbl}
\usepackage{multirow} 
\usepackage{enumitem}
\usepackage{bbm}
\usepackage{bm}


\definecolor{cvprblue}{rgb}{0.21,0.49,0.74}
\usepackage[pagebackref,breaklinks,colorlinks,allcolors=cvprblue]{hyperref}

\def\paperID{11314} 
\def\confName{CVPR}
\def\confYear{2025}

\title{Understanding and Mitigating the Bias in Sample Selection\\ for Learning with Noisy Labels}

\author{
Qi Wei\textsuperscript{\rm 1},\,
Lei Feng\textsuperscript{\rm 2}\thanks{Corresponding author},\,
Haobo Wang\textsuperscript{\rm 3},\,
Bo An\textsuperscript{\rm 1} \\
\textsuperscript{\rm 1}College of Computing and Data Science, Nanyang Technological University \\
\textsuperscript{\rm 2}Information Systems Technology and Design, Singapore University of Technology and Design \\
\textsuperscript{\rm 3}School of Software, Zhejiang University \\
}

\begin{document}
\maketitle
\input{0_abstract}    
\section{Introduction}\label{sec1_intro}\input{1_intro}

\section{Understanding the Bias in Sample Selection}\input{3_analysis}

\section{Methodology}\label{sec4_method}\input{4_method}
\section{Experiments}\label{sec5_experiment}\input{5_experiment}

\section{Related Work}\input{2_related_new}

\section{Conclusion}\label{sec6_conclusion}\input{6_conclusion}

{
    \small
    \bibliographystyle{ieeenat_fullname}
    \bibliography{main}
}

\input{appendix}

\end{document}

%% file: 0_abstract.tex
\begin{abstract}
Learning with noisy labels aims to ensure model generalization given a label-corrupted training set.
The sample selection strategy achieves promising performance by selecting a label-reliable subset for model training.
In this paper, we empirically reveal that existing sample selection methods suffer from both data and training bias, which are represented in practice as imbalanced selected sets and accumulation errors.
However, only the training bias was handled in previous studies. 
To address this limitation, we propose a noIse-Tolerant Expert Model (ITEM) for debiased learning in sample selection.
Specifically, to mitigate the training bias, we design a robust Mixture-of-Expert network that conducts selection and learning on different layers. 
Compared with the prevailing double-branch network, our network performs better on both selection and prediction by ensembling multiple experts while training with fewer parameters. 
Meanwhile, to mitigate the data bias, we propose a weighted sampling strategy that assigns larger sampling weights to classes with smaller frequencies.
Using MixUp, the model is trained on a mixture of two batches: one sampled by a weighted sampler and the other by a regular sampler, which mitigates the effect of the imbalanced training set while avoiding sparse representations that are easily caused by sampling strategies.
Extensive experiments on seven noisy benchmarks and analyses demonstrate the effectiveness of ITEM. 
The code is released at \url{https://github.com/1998v7/ITEM}.
\end{abstract}

%% file: 1_intro.tex
The remarkable generalization capability of deep neural networks (DNNs) is achieved through training on a large-scale dataset. However, existing training sets are usually collected by online queries \cite{blum2003noise}, crowdsourcing \cite{yan2014learning}, and manual annotations, which could inevitably incur wrong (or noisy) labels \cite{wei2023fine}. 
Since DNNs exhibit vulnerability to such low-quality training sets \cite{zhang2017understanding}, training on the label-corrupted set presents a great challenge for the modern machine-learning community. To mitigate the adverse effect brought by noisy samples, learning with noisy labels (LNL) \cite{wei2020combating,xia2021sample} is important, contributing to improvements of the model's generalization on practical applications.

\begin{figure}[tp]
    \centering
    \includegraphics[width=0.75\linewidth]{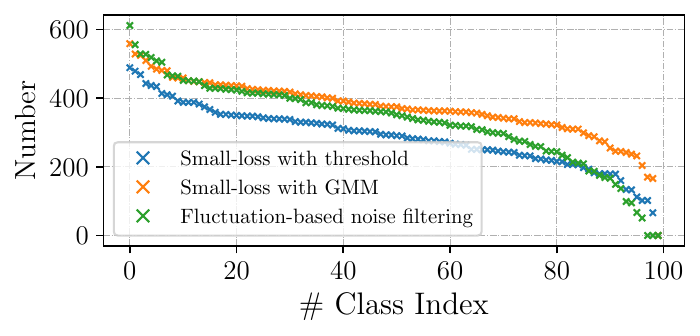}  
    \vspace{-3mm}
    \caption{\small \textbf{Existing selection criteria always lead to an imbalanced training set}, termed as the data bias. A ResNet-34 is trained on CIFAR-100N. We visualize the class distribution of the selected set, given three typical selection criteria. The quantity of class-level samples in the last epoch is counted, while the index of classes is sorted. More results can be found in Appendix A.}
    \label{fig:imbalanced_intro}
    \vspace{-3mm}
\end{figure}

\begin{figure*}[t]
\centering
\subfloat[\footnotesize A corrupted selection tendency]{
\begin{minipage}[b]{0.6\linewidth}
\includegraphics[width=\linewidth]{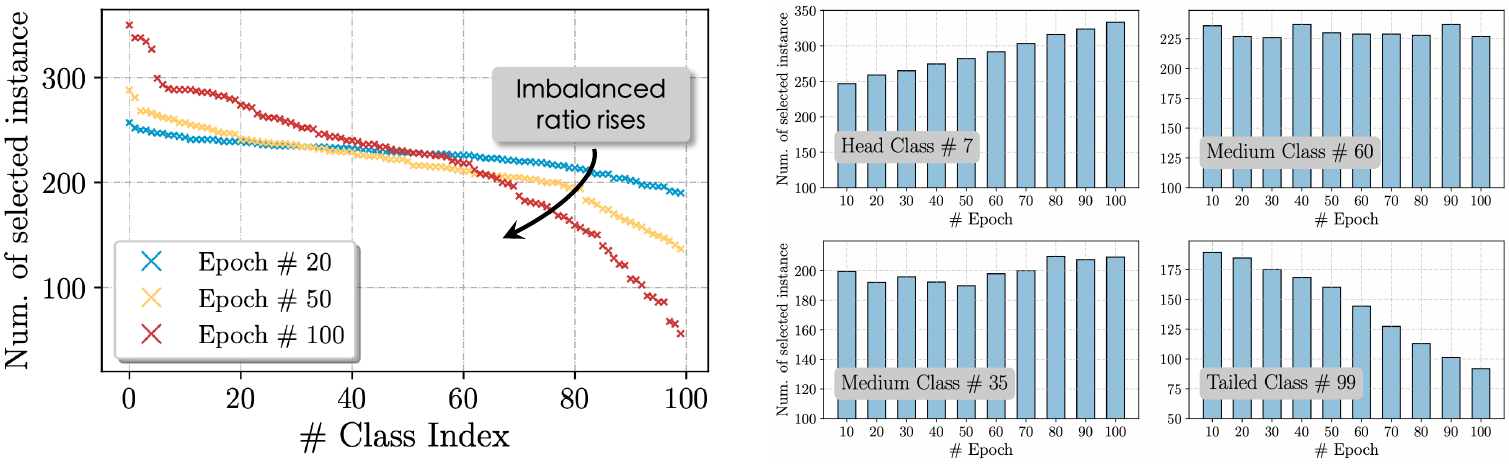}
\end{minipage}}
\hfil
\subfloat[\footnotesize Class-level F-score]{
\begin{minipage}[b]{0.31\linewidth}
\includegraphics[width=\linewidth]{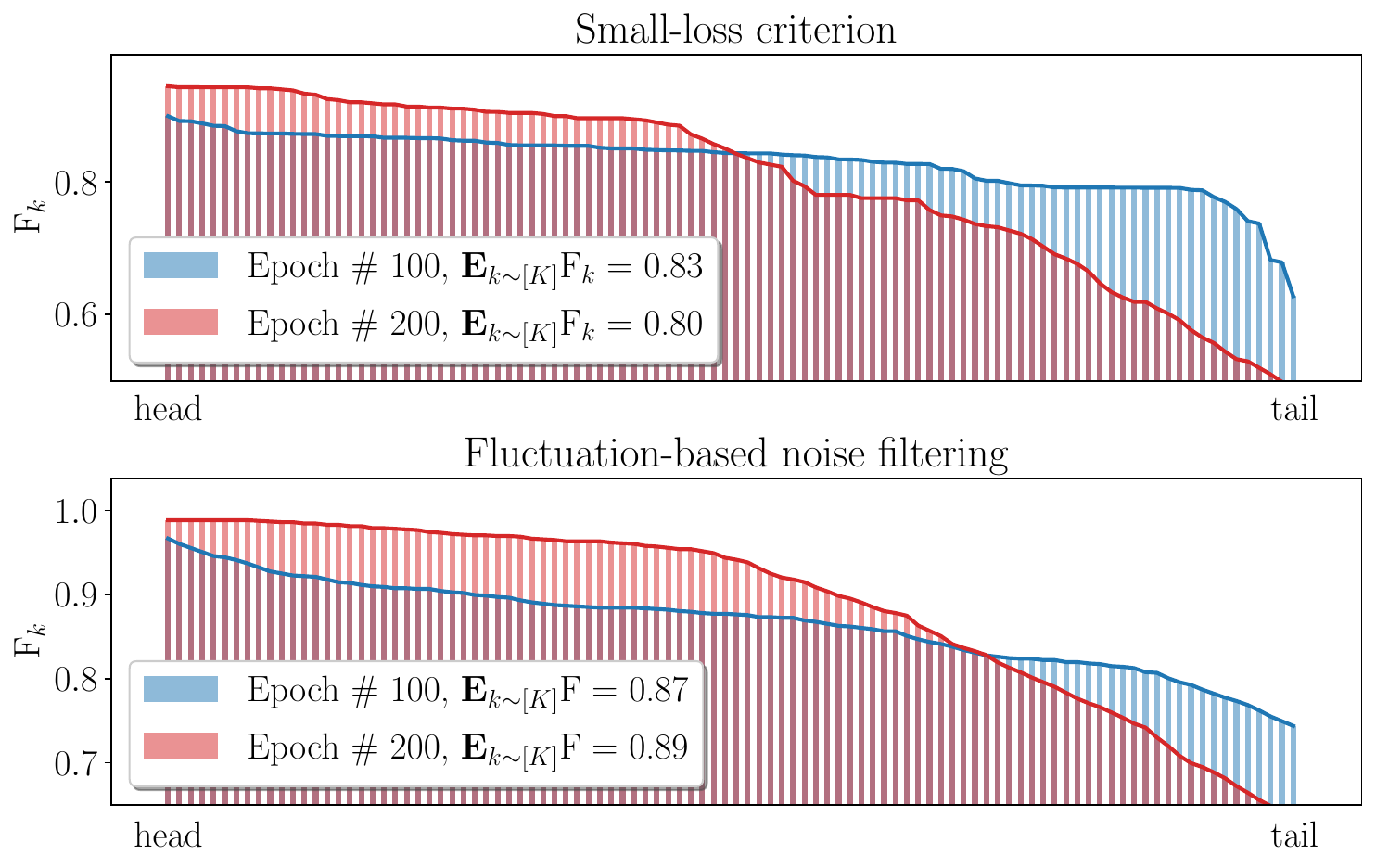}
\end{minipage}}
\vspace{-2mm}
\caption{\small We train ResNet-18 on Sym. 60\% CIFAR-100 with small-loss \cite{han2018co} and fluctuation-based noise filtering \cite{wei2022self}. (\textbf{a}) Results on small-loss. \emph{Left}: the class distribution of three training stages. \emph{Right}: dynamic numbers of four representative categories. (\textbf{b}) Results on two selection criteria. Visualization of the class-level selection performance (F-score) $\text{\sffamily F}_k$ in 100- and 200-th epoch.}
\label{fig:vis}
\vspace{-2mm}
\end{figure*}

\emph{Sample selection} \cite{han2018co,li2020dividemix, wei2022self}, a prevailing strategy for LNL, achieves considerable performance in mitigating the effects of noisy labels \cite{xu2019l_dmi} by carefully selecting clean samples from the label-corrupted training set.
The performance of sample selection approaches is largely decided by the selection criteria, which can be roughly categorized into two sides:
1) \textit{small-loss} based strategies \cite{wei2020combating,han2018co,li2020dividemix,yu2019does,liu2020early,bai2021me}, which are motivated by the memorization effect \cite{arpit2017closer} that DNNs learn simple patterns shared by majority examples before fitting the noise. 
Hence, the samples with small losses in the early learning stage can normally be taken as clean samples.
2) \textit{fluctuation} based strategies \cite{zhou2020robust,wei2022self,yuan2023late,xia2021sample}, which are motivated by the observation that DNNs easily give inconsistent prediction results for noisy samples. 
These methods \cite{wei2022self,yuan2023late} normally consider an instance incorrectly labeled if its prediction results exhibit alterations within two consecutive training rounds or during a specified subsequent period.

Despite the validated effectiveness, there is a common view shared by existing sample selection methods that the sample selection processes are only influenced by a training bias, i.e., the accumulated error.
In this paper, we, for the first time, show that by extensive experiments, there exists another type of bias, i.e., the \emph{data bias}, which is mainly caused by imbalanced data distribution. 
To be concrete, we split the bias in a selection-based learning framework into the training bias, which inherently exists in a self-training manner, and the data bias, which indicates the selected set tends to be class-imbalanced (see Figure \ref{fig:imbalanced_intro}).
Moreover, we customize a class-level F-score and do extensive experiments in the next section, providing three nontrivial findings that reveal the adverse effect of the data bias.
These observations motivate us to consider a novel selection-based learning framework that can simultaneously address training and data bias.

To this end, we propose a noIse-Tolerant Expert Model (ITEM), consisting of a noise-robust multi-experts structure and a weighted sampling strategy. 
\begin{itemize}
\setlength\itemsep{0mm}
    \item Firstly, motivated by the idea of ensemble learning \cite{dong2020survey} and the Mixture-of-Expert model \cite{kotsiantis2006machine}, we design a robust network structure integrating the classifier with multiple experts. Our network is more robust compared with existing robust networks like double-branch network \cite{han2018co} for the following reasons, 1) 
    in our structure, the classifier learning and selection are disentangled, with experts selecting clean samples for the classifier's training, naturally mitigating the issue of error accumulation. 2) Ensembling selection results from different experts reduces harmful interference from incompatible patterns (\textit{e.g.}, poor selection made by individual experts), leading to a cleaner selected set.

    \item Secondly, to mitigate the implicit data bias during selection, we propose a weighted resampling strategy named mixed sampling. By calculating the quantity of each class in the selected set, we assign larger weights to tail classes through a mapping function during sampling. Finanlly, introducing the MixUp  \cite{zhang2017mixup}, our network is trained on a mixed batch that combines two batches based on a regular and a weighted sampler, fulfilling class-balanced learning while avoiding the issue of sparse representations.
\end{itemize}

The promising performance of ITEM is verified on seven noisy benchmarks. Extensive ablation studies and analyses also demonstrate the effectiveness of each component.

%% file: 3_analysis.tex

In this section, we first introduce the formal problem setting of learning with noisy labels (LNL) and its learning goal. Then, we conduct experiments to deeply analyze where the bias of sample selection in LNL comes from.

Assume $\mathcal{X}$ is the feature space and $\mathcal{Y} = \{1,2,\ldots, K\}$ is the label space. Suppose the training set is denoted by ${\tilde D} = \{({\bm x}_i, y_i)\}_{i \in [N]}$, where $[N] = \{1,2,\ldots,N\}$ is the set of indices. Since the annotator may give wrong labels in practice \cite{krizhevsky2012imagenet}, the learner thus can only observe a label-corrupted set ${D_N} = \{({\bm x}_i, {\tilde y}_i)\}_{i \in [N]}$ with noisy labels.

As a prevailing strategy for LNL, sample selection \cite{wei2022self,li2023disc,li2020dividemix,xia2021sample} aims to  progressively select a reliable subset ${D_M} \subseteq {D_N}$ ($M<N$) and feed ${D_M}$ to the classifier for training. We introduce $v_i \in \{0,1\}$ to indicate whether the $i$-th instance is selected ($v_i = 1$) or not ($v_i = 0$). Generally, the performance of a selection-based method can be reflected by the deviation between the actual selected samples and the total clean samples, which can be measured by the F-score $\text{\sffamily F}$, where $\text{\sffamily F} = \frac{2 \cdot \text{\sffamily P} \cdot \text{\sffamily R}}{\text{\sffamily P} + \text{\sffamily R}}$, $\text{\sffamily P}$ and \text{\sffamily R} denote selection precision and recall, respectively. In this paper, we individually calculate each class's F-score for further analysis. Specifically, the class-level F-score for class $k$ is written as 
\begin{equation}\nonumber
    \text{\sffamily F}_k = \frac{2 \cdot \text{\sffamily P}_k \cdot \text{\sffamily R}_k}{\text{\sffamily P}_k + \text{\sffamily R}_k}, \,\, \text{where} \\
    \left\{
    \begin{array}{lr}
              \text{\sffamily P}_k = \frac{\sum \nolimits_{i\in[N]} \mathbbm{1}(v_i=1, y_i=\Tilde{y}_i=k)}{\sum \nolimits_{i\in[N]} \mathbbm{1}(v_i=1, \Tilde{y}_i=k)}, & \\
              \text{\sffamily R}_k = \frac{\sum \nolimits_{i\in[N]} \mathbbm{1}(v_i=1, y_i=\Tilde{y}_i=k)}{\sum \nolimits_{i\in[N]} \mathbbm{1}(y_i=\Tilde{y}_i=k)}. &
        \end{array}
        \right.
\end{equation}

\begin{figure}[t]
    \centering
    \includegraphics[width=0.99\linewidth]{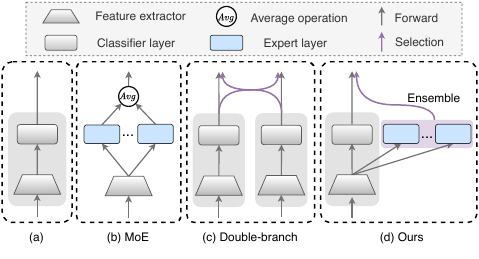}
    \caption{\small \textbf{Comparisons of different architectures}. 
    \textbf{(a)} \textit{Typical classification network}, which consists of a feature extractor and a classifier layer. 
    \textbf{(b)} \textit{Mixture-of-experts (MoE)} \cite{masoudnia2014mixture}, a set of experts jointly gives the predicted label for the input. 
    \textbf{(c)} \textit{Double-branch robust structure}, the network is trained on a selected set that is considered clean by another network.
    \textbf{(d)} \textit{Ours}, a mixture-of-experts module, is integrated into the classification network, which works for robust selection as well as prediction ensemble. 
    }
    \vspace{-2mm}
    \label{fig:framework}
\end{figure}

By analyzing the selection performance under different training conditions, we have several nontrivial findings.

\begin{itemize}
    \item \textit{Current selection criteria easily result in an imbalanced selection subset}. As shown in Figure \ref{fig:imbalanced_intro}, we can see that both of the two adopted selection strategies incur the imbalanced data distribution under two noisy conditions. The reason is that both of the two selection criteria rely on model performance. However, the model has different capacities for different classes. For those classes with indistinguishable characteristics (i.e., rare classes), the model tends to produce a large loss or inconsistent results and further discards those samples.

    \item \textit{Sample selection, conducted as self-training, will exacerbate the imbalanced ratio during the training process}. As shown in Figure \ref{fig:vis} (a), the imbalanced ratio of the selected set increases as the training process proceeds, i.e., the number of selected instances increases in head classes and decreases in tail classes (see the right part of Figure \ref{fig:vis}). The intuitive reason is that the model's performance on tail classes hardly improves due to the limited number of available samples for training, which further degrades the effectiveness of selection criteria on these classes.    

    \item \textit{The selection performance is inherently influenced by the imbalanced class distribution.} As shown in Figure \ref{fig:vis} (b), the selection performance of both selection criteria increases in head classes but decreases in tail classes. The reason is that the selected errors are relatively small for categories with higher F-scores. Hence, the training of these classes would further improve the model's accuracy and selection performance. However, for tail classes with lower F-score, the self-training mechanism would further increase the selection error and thus fail to select instances in the follow-up phase correctly.
\end{itemize}
Based on these observations, 
we can deduce that the classifier under a sample selection framework implicitly suffers from a data bias, which is, however, rarely studied in previous literature. Motivated by this, we aim to design a model that confronts not only accumulated error but also the imbalanced selected set.

%% file: 4_method.tex
\textbf{Preliminaries.} In LNL, we are given a label-corrupted training set ${D_N} = \{({\bm x}_i, {\tilde y}_i)\}_{i \in [N]}$ of $N$ samples. A classifier with learnable parameters $f_{\bm{\theta}}$ is a function that maps from the input space $\mathcal{X}$ to the label space $f:\mathcal{X} \rightarrow \mathbb{R}^K$. In multi-class classification, we always update the parameter $\bm{\theta}$ by minimizing the following empirical risk:
\begin{equation}
    \hat{R}(f) = \frac{1}{N} \sum\nolimits_{i=1}^{N}\mathcal{L}(f(\bm{x}_i,\bm{\theta}),  {\tilde y}_i),
\end{equation}
where $\mathcal{L}(\cdot)$ is the given loss function, e.g., Softmax Cross-Entropy (CE) loss. In sample selection strategy, previous works mainly focus on designing high-effectively selection criteria. Here, we give a universal training objective on the selected clean set, which is written as
{
\begin{align}
    \hat{R}_{\text{clean}} (f)  &= \frac{1}{N} \sum\nolimits_{i=1}^{N}
\big{[} \mathrm{Criterion}(f, \bm{x}_i, {\tilde y}_i) \cdot \mathcal{L}(f(\bm{x}_i,\bm{\theta}), {\tilde y}_i) \big{]}, \quad \nonumber \\
    & \text{where} \,\,
    \mathrm{Criterion}(\cdot) = 
    \left\{
        \begin{array}{lr}
              1, \,\, \text{If selected}, & \\
              0. &
        \end{array}
    \right.
\end{align}
}

\noindent In this unified formula, \emph{the classifier intended to be optimized is consistent with the classifier serving as the noise filter during the selection phase}, which easily leads to the training bias. Empirical results in the section above show that directly optimizing ${\hat R}_{\text{clean}} (f)$ with an existing selection criterion is always biased, further resulting in inferior generalization performance.

\noindent\textbf{Overview.} We propose to fulfill debiased learning by handling both the training and data bias. First, to solve the training bias, we design a novel network architecture called noIse-Tolreant Expert Model (ITEM). Compared with the widely applied double-branch network, our proposal exhibits greater robustness to noisy labels. Second, to solve the data bias, we propose a weighted resampling strategy that can mitigate the side-effect caused by the class-imbalanced set $D_{\text{clean}}$ while avoiding sparse representations.

\subsection{ITEM: noIse-Tolerant Expert Model}\label{sec:41_model}
The training bias in sample selection arises from the self-training approach. Previous works \cite{han2018co,li2020dividemix} always resort to the double-branch network to confront this problem. Concretely, the network is trained on the clean set that is selected by the other network. Motivated by Mixture-of-experts \cite{rokach2010ensemble,masoudnia2014mixture}, we propose a robust architecture that integrates multiple experts into the classifier, which independently conducts classification and sample selection.

Specifically, compared with a normal classification network with a classifier layer $f$, our network ITEM contains $f$ and a set of additional expert layers $\{g^1,...,g^m\}$ with size $m$. Each expert layer has the same dimension as the classifier layer, \emph{i.e.}, $\Vert f\Vert=\Vert g\Vert$. For robust selection, we propose to conduct existing criteria on expert layers instead of the classifier layer. Consequently, the training objective in our framework can be summarized as
\begin{align}\label{eq:our_coupledxx}
    \hat{R}_{\text{clean}} (f)  = \frac{1}{N} \sum\nolimits_{i=1}^{N}
     \big{[}\mathrm{Criterion}&(\{g^1,...,g^m\}, \bm{x}_i, {\tilde y}_i) \cdot \nonumber\\
     &\mathcal{L}(f(\bm{x}_i,\bm{\theta}), {\tilde y}_i)\big{]}.
\end{align}
Intuitively, the independent selection phase prevents the classifier layer from selecting noisy samples and updating its parameters based on them, significantly mitigating the training bias. In practice, a voting strategy is applied to ensemble the selection results of $m$ experts, where the sample selected by all experts is considered clean. It is noteworthy that the selection criterion $\mathrm{Criterion}(\cdot)$ is not restricted to a specific form (e.g., small-loss based or fluctuation-based), which highlights the applicability of this expert model as a backbone in sample selection.

\noindent\textbf{Advantages of the MoE structure.} We compare different network structures in LNL in Figure \ref{fig:framework}. Compared with the prevailing double-branch networks, which have proved to be noise-robust in previous works \cite{han2018co,li2020dividemix}, our architecture offers two advantages, including 
1) \emph{Greater robustness to noisy labels}. Each expert exhibits different capabilities during training, resulting in a better inductive bias for both selection and prediction. Specifically, by ensembling the selections from different experts, the MoE minimizes negative interference from incompatible patterns (e.g., poor selection made by individual experts), thereby enhancing the framework's robustness against noisy labels.
2) \emph{Resource-friendly training}. In contrast to the feature extractor's huge parameters, the last layer's parameters for classification are significantly fewer. The additional learnable parameters in our model are less than 2\% of those in the double-branch network. Thus, our network is more friendly to GPU and computer memory.

\subsection{Beta-Resampling}

\begin{figure}[tp]
    \centering
    \includegraphics[width=0.6\linewidth]{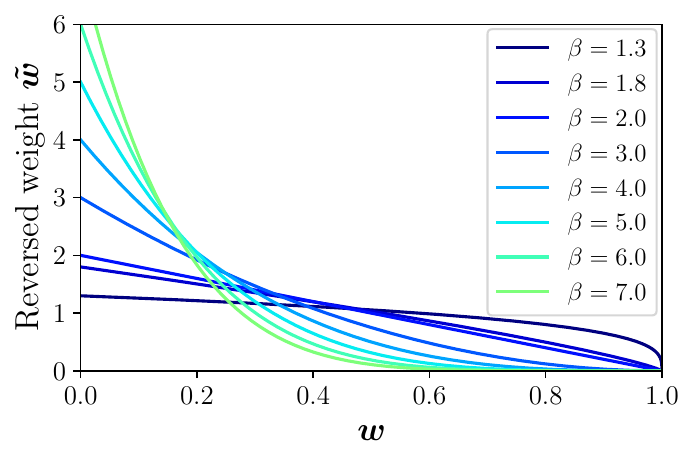}  \vspace{-3mm}
    \caption{\small Mapping function $\mathcal{S}^{\beta}(\cdot)$ with different values of $\beta$.}
    \vspace{-2mm}
    \label{fig:mapping_func}
\end{figure}

To overcome the data bias, we propose a mixed beta-sampling strategy that promotes the sampling frequency for tail classes in the selected clean set when sampling. 
Thus, the model can learn from more knowledge from the tailed classes, contributing to mitigating the data bias.

Specifically, suppose a clean set ${D}_{\text{clean}}$ is selected from the ${D_N}$, where ${D}_{\text{clean}} \subseteq {D_N}$. 
The set of samples labeled as class $k$ in ${D}_{\text{clean}}$ is ${D}_{\text{clean}}^k$. By introducing L1 normalization, a vector of class frequencies $\bm{v}$ can be obtained, 
\begin{equation}\label{eq:v}
\bm{v} = [w_1,w_2,...w_K], \, \text{where} \,\, w_k = \frac{\Vert {D}_{\text{clean}}^k \Vert}{\Vert {D}_{\text{clean}} \Vert}.
\end{equation}
Here, we can regard the class frequency as the sampling weight, \textit{i.e.}, the class with a larger frequency will be sampled with a greater weight during the data loading phase.

To mitigate the data bias, which leads the model to learn from the majority classes mainly, we propose a mapping function that maps the lower frequency to a higher weight when data sampling. Concretely, this mapping function stems from the probability density function (PDF) of a Beta distribution. Normally, the Beta distribution's PDF is controlled by two parameters $\alpha, \beta$. To satisfy both 1) the input interval of ${w}_k$ belongs to $[0,1]$ and 2) the function contains the monotonically decreasing property, we thus fix $\alpha=1$ and adjust $\beta$ for different slopes. Given a weight $w_i$, the reversed weight ${\tilde w}_i$ is written as 
\begin{equation}
\label{eq:mapping}
    {\tilde w_i} = \mathcal{S}^{\beta}(w_i) =
    \frac{1}{\text{B}(1,\beta)}\,(1-w_i)^{\beta-1}, 
\end{equation}
where $\beta$ is a hyper-parameter and B$(\cdot)$ denotes a beta function written as B$(1, \beta) = \int_{0}^{1} (1-t)^{\beta-1} dt$. 

The mapping function $\mathcal{S}^{\beta}(\cdot)$ for varying values of $\beta$ is illustrated in Figure \ref{fig:mapping_func}. It can be observed that as the input weight increases, the output value decreases. Consequently, this mapping function enables the transformation of class frequencies in Eq. \ref{eq:v} into a tail-focused weighted vector ${\bm{\tilde{v}}} = [\tilde{w}_1, \ldots, \tilde{w}_K]$. By weighted sampling with ${\bm{\tilde v}}$, a tail-classes focused training batch $B_{{\bm{\tilde v}}}$ can be obtained. Training the model on $B_{{\bm{\tilde v}}}$ can effectively mitigate the data bias.

\begin{table*}[t]
\centering
\caption{\small Test accuracy (mean$\pm$std) of methods using ResNet-18/34 on CIFAR-10/100. Note that $\dagger, \ddagger$ and $\sharp$ denote three selection criteria, \textit{i.e.}, small-loss selection with loss threshold \cite{han2018co}, small-loss selection with Gaussian Mixture Model \cite{li2020dividemix}, and fluctuation-based noise filtering \cite{wei2022self}. \textbf{Bold} values denote the best the second best performance.
}
\vspace{-2mm}
\label{tab:results_onCIFAR}
\resizebox{0.95\textwidth}{!}{
\setlength{\tabcolsep}{4mm}{
\begin{tabular}{l|lcccc|c}
\toprule[0.9pt]
&Methods & Sym. 20\% & Sym. 40\% & Inst. 20\% & Inst. 40\%   & \emph{Avg.} \\ \midrule[0.9pt]
\multirow{14}{*}{\rotatebox{90}{CIFAR-10}} 
&Cross-Entropy & 85.00 $\pm$ 0.43\% & 79.59 $\pm$ 1.31\% & 85.92 $\pm$ 1.09\% & 79.91 $\pm$ 1.41\% & 82.61 \\
&JoCoR \cite{wei2020combating} & 88.69 $\pm$ 0.19\% & 85.44 $\pm$ 0.29\% & 87.31 $\pm$ 0.27\% & 82.49 $\pm$ 0.57\%  &85.98\\
&Joint Optim \cite{tanaka2018joint} & 89.70 $\pm$ 0.36\% & 87.79 $\pm$ 0.20\% & 89.69 $\pm$ 0.42\% & 82.62 $\pm$ 0.57\% & 87.45\\
&CDR \cite{xia2020robust} & 89.68 $\pm$ 0.38\% & 86.13 $\pm$ 0.44\% & 90.24 $\pm$ 0.39\% & 83.07 $\pm$ 1.33\%  & 87.28\\
&Me-Momentum \cite{bai2021me} & {91.44 $\pm$ 0.33\%} & 88.39 $\pm$ 0.34\% & 90.86 $\pm$ 0.21\% & 86.66 $\pm$ 0.91\% & 89.34 \\
&PES \cite{bai2021understanding}   & 92.38 $\pm$ 0.41\%  & 87.45 $\pm$ 0.34\% & 92.69 $\pm$ 0.42\%   & 89.73 $\pm$ 0.51\%  & 90.56\\
&Late Stopping \cite{yuan2023late} & 91.06 $\pm$ 0.22\% & {88.92 $\pm$ 0.38\%} & {91.08 $\pm$ 0.23\%} & {87.41 $\pm$ 0.38\%} & 89.62\\  \cmidrule{2-7}
&$\dagger$ Co-teaching \cite{han2018co} & 87.16 $\pm$ 0.52\% & 83.59 $\pm$ 0.28\% & 86.54 $\pm$ 0.11\% & 80.98 $\pm$ 0.39\% & 84.56\\
&$\dagger$ ITEM {\scriptsize (Ours)} & 93.79 $\pm$ 0.14\% & 90.83 $\pm$ 0.19\% & 93.52 $\pm$ 0.14\% & 91.09 $\pm$ 0.18\% & 92.31\\ 
&$\ddagger$ ITEM {\scriptsize (Ours)} & \textbf{95.01 $\pm$ 0.21\%}  & \textbf{93.10 $\pm$ 0.20\%} & \textbf{95.18 $\pm$ 0.19\%} & \textbf{93.65 $\pm$ 0.12\%} & \textbf{94.24} \\ \cmidrule{2-7}
&$\sharp$ SFT \cite{wei2022self} & 92.57 $\pm$ 0.32\%  & 89.54 $\pm$ 0.27\% & 91.41 $\pm$ 0.32\%  & 89.97 $\pm$ 0.49\% & 90.87\\ 
&$\sharp$ ITEM {\scriptsize (Ours)} & \textbf{95.26 $\pm$ 0.23\%}  & \textbf{92.81 $\pm$ 0.20\%} & \textbf{95.80 $\pm$ 0.18\%} & \textbf{93.13 $\pm$ 0.29\%} & \textbf{94.25}\\ \midrule[0.9pt]
\multirow{14}{*}{\rotatebox{90}{CIFAR-100}} 
&Cross-Entropy & 57.59 $\pm$ 2.55\% & 45.74 $\pm$ 2.61\% & 59.85 $\pm$ 1.56\% & 43.74 $\pm$ 1.54\% & 51.73 \\
&JoCoR \cite{wei2020combating} & 64.17 $\pm$ 0.19\% & 55.97 $\pm$ 0.46\% & 61.98 $\pm$ 0.39\% & 50.59 $\pm$ 0.71\% & 58.17\\
&Joint Optim \cite{tanaka2018joint} & 64.55 $\pm$ 0.38\% & 57.97 $\pm$ 0.67\% & 65.15 $\pm$ 0.31\% & 55.57 $\pm$ 0.41\% & 60.81\\
&CDR \cite{xia2020robust} & 66.52 $\pm$ 0.24\% & 60.18 $\pm$ 0.22\% & 67.06 $\pm$ 0.50\% & 56.86 $\pm$ 0.62\% & 62.65\\
&Me-Momentum \cite{bai2021me} & 68.03 $\pm$ 0.53\% & 63.48 $\pm$ 0.72\% & 68.11 $\pm$ 0.57\% & 58.38 $\pm$ 1.28\%  & 64.50\\
&PES \cite{bai2021understanding}  & 68.89 $\pm$ 0.41\%  & 64.90 $\pm$ 0.57\%  & 70.49 $\pm$ 0.72\%  & 65.68 $\pm$ 0.44\%    & 67.49\\
&Late Stopping \cite{yuan2023late}  & {68.67 $\pm$ 0.67\%} & {64.10 $\pm$ 0.40\%} & {68.59 $\pm$ 0.70\%}  & {59.28 $\pm$ 0.46\%} & 65.16\\  \cmidrule{2-7}
&$\dagger$ Co-teaching \cite{han2018co} & 59.28 $\pm$ 0.47\% & 51.60 $\pm$ 0.49\% & 57.24 $\pm$ 0.69\% & 45.69 $\pm$ 0.99\% & 53.45\\
&$\dagger$ ITEM {\scriptsize (Ours)} & 72.42 $\pm$ 0.17\% & 71.96 $\pm$ 0.24\% & 73.61 $\pm$ 0.16\% & 69.90 $\pm$ 0.30\% & 71.97\\ 
&$\ddagger$ ITEM {\scriptsize (Ours)} & \textbf{78.20 $\pm$ 0.09\%} & \textbf{75.27 $\pm$ 0.20\%} & \textbf{77.91 $\pm$ 0.14\%} & \textbf{70.69 $\pm$ 0.31\%} & \textbf{75.51}\\ \cmidrule{2-7}
&$\sharp$ SFT \cite{wei2022self}  & 71.98 $\pm$ 0.26\%   & 69.72 $\pm$ 0.31\%   & 71.83 $\pm$ 0.42\%   & 69.91 $\pm$ 0.54\% & 70.86 \\
&$\sharp$ ITEM {\scriptsize (Ours)} & \textbf{77.19 $\pm$ 0.13\%} & \textbf{74.90 $\pm$ 0.24\%} & \textbf{76.91 $\pm$ 0.23\%} & \textbf{71.44 $\pm$ 0.29\% }& \textbf{75.11}\\
\bottomrule[0.9pt]  
\end{tabular}
}
}
\end{table*}


\subsection{Stochastic Classifier Training}
Since the noise filter and the optimizer are disentangled in Eq. (\ref{eq:our_coupledxx}), largely mitigating the training bias, the parameter update of expert layers is unreachable.
Considering almost all existing selection criteria rely on the model's performance, dynamically updating the parameters of expert layers is necessary.
To fulfill this goal, we propose a stochastic training strategy, which randomly assigns a layer from the set of 
$\mathbbm{F} = \{f, g^1,...,g^m\}$ as the classifier layer $f^c$ and the rest are expert layers. Hence, Eq. (\ref{eq:our_coupledxx}) can be rewritten as
\begin{align}\label{eq:our_coupled2}
     \hat{R}_{\text{clean}} (f^c) =  \frac{1}{N} \sum\nolimits_{i=1}^{N}
     \big{[}\mathrm{Criterion}&(\mathbbm{F} \backslash f^c, \bm{x}_i, {\tilde y}_i) \cdot \nonumber\\
     &\mathcal{L}(f^c(\bm{x}_i,\bm{\theta}), {\tilde y}_i) \big{]}.
\end{align}
Since the aim-to-optimized layer $f^c$ is initialized from the set $\mathbbm{F}$ in each iteration, the performance of each layer can be ensured under sufficient training processes.

Despite a highly integrated framework in Eq. (\ref{eq:our_coupled2}) proposed, the training manner is practically decoupled into three stages that are proceeding iteratively, 
1) select clean samples via the preset criterion based on the ensemble result of all experts, 2) randomly select an optimization layer, and 3) train this layer on the selected set.

Considering training on the tail-focused training batch directly would result in the sparse representation of head classes \cite{zhou2020bbn}, we separately conduct two times weighted samples (according to the weighted and reversed weighted vector ${\bm{v, \tilde v}}$) in each data-loading phase, which focus on head and tail classes, respectively. By leveraging the MixUp strategy \cite{zhang2017mixup}, the model can learn a more representative feature extractor on the mixed data. Specifically, in each training iteration, we respectively sample a head-focused batch $B_{\bm{v}} = \{(\bm{x}_i, {\tilde y}_i)\}_{i=1}^b$ and a tail-focused batch $ B_{{\bm{\tilde v}}} = \{(\bm{x'}_i, {\tilde y'}_i)\}_{i=1}^b$ from ${\tilde D}_{\text{clean}}$ according to ${\bm{\tilde v}}$, where $b$ denotes the size of the mini-batch. 
By randomly selecting a layer from $\mathbbm{F}$ as the optimized-classifier layer $f^c$, the parameter's update can be achieved by minimizing
\begin{equation}\label{eq:trainLoss}
    L^{\mathrm{train}} 
    = \frac{1}{b} \sum\nolimits_{i=1}^b
    \mathcal{L} \big{(}f^c \big{(}\mathrm{MixUp}(\bm{x}_i, \bm{x'}_i) \big{)}, \, \mathrm{MixUp}({\tilde y}_i, {\tilde y}'_i)  \big{)},
\end{equation}
where $\mathrm{MixUp}(,)$ is represented as $\mathrm{MixUp}(a,b)=\gamma \cdot a + (1-\gamma) \cdot b$ and $\gamma$ is a trade-off coefficient randomly sampled from a beta distribution $\texttt{beta}(\delta,\delta)$.
The algorithm flowchart of ITEM is shown in Algorithm \ref{alg:algorithm} (see Appdx).


%% file: 5_experiment.tex

\textbf{Datasets.} 
We assess the performance of \textit{ITEM} on two noise-synthetic datasets CIFAR-10 and CIFAR-100 \cite{krizhevsky2012imagenet}, two human-annotated datasets CIFAR10N and CIFAR100N \cite{wei2021learning}, and three real-world noisy benchmarks including Clothing-1M \cite{xiao2015learning}, Food-101N \cite{lee2018cleannet}, and Webvision \cite{li2017webvision}.

\noindent\textbf{Baselines.}
On CIFAR-10 and CIFAR-100, we compare ITEM with prevailing methods that can be roughly divided into three parts, including 1) \textit{small-loss based selection}, JoCoR \cite{wei2020combating}, and Me-Momentum \cite{bai2021me}, 2) \textit{fluctuation-based selection}, SFT \cite{wei2022self} and Late Stopping \cite{yuan2023late}, 3) others, Cross-Entropy, Joint Optim \cite{tanaka2018joint}, and PES \cite{bai2021understanding}. Reported results in this paper are collected from SFT and Late Stopping. More  information is shown in Appdx.


\begin{table}[t]
\centering
\caption{\small Test accuracy ($\%$) of prevailing methods using ResNet-34 on CIFAR10N and CIFAR100N. Note that \ding{51} and \ding{55} indicate whether a \emph{semi-supervised framework} is used or not.}
\vspace{-3mm}
\resizebox{0.48\textwidth}{!}{
\setlength{\tabcolsep}{1.3mm}{
\begin{tabular}{l|c|ccccc}
\toprule[0.9pt]
\multicolumn{2}{c|}{\multirow{2}{*}{Methods}} & \multicolumn{4}{c}{CIFAR10N} & \multirow{2}{*}{CIFAR100N} \\
\multicolumn{1}{c}{} & \multicolumn{1}{c|}{} & {Worst} & {R1}& {R2} & {R3} &             \\ \midrule
Co-teaching+ \cite{yu2019does}                   & \ding{55}   & 83.26     & 89.70 & 89.47 & 89.54  & 60.37                  \\
Peer Loss \cite{liu2020peer}                      & \ding{55}   & 82.00     & 89.06 & 88.76 & 88.57  & 57.59                   \\
CAL \cite{zhu2021second}                            & \ding{55}   & 85.36     & 90.93 & 90.75 & 90.74  & 61.73                  \\
Late Stoping \cite{yuan2023late}                    & \ding{55}   & 85.27     & -     & -     & -      & -                       \\
$\ddagger$ ITEM  & \ding{55}   & \textbf{91.15}     & \textbf{95.12} & \textbf{95.03} & \textbf{95.17} & \textbf{69.47}  \\ \midrule
DivideMix \cite{li2020dividemix}                      & \ding{51}   & 92.56  & 95.16 & 95.23 & 95.21 & 71.13                     \\
ELR+ \cite{liu2020early}                           & \ding{51}   & 91.09  & 94.43 & 94.20 & 94.34 & 66.72                     \\
CORES*  \cite{cheng2020learning}                        & \ding{51}   & 91.66  & 94.45 & 94.88 & 94.74 & 61.15                     \\
DPC \cite{zong2024dirichlet}     & \ding{51} & \textbf{93.82} & 95.97     & 95.92     & 95.90 & 71.42 \\
$\ddagger$ ITEM* & \ding{51}  & 93.14  & \textbf{96.08} & \textbf{96.54} & \textbf{96.71} & \textbf{72.40}  \\ 
\bottomrule[0.9pt]  
\end{tabular}
}}
\label{tab:cifar10N}
\end{table}

\noindent\textbf{Implementation details.} Our code implements utilize Pytorch 1.9.0 and all experiments are run on a single RTX 4090 GPU. 
We keep the convention from \cite{wei2022self,yuan2023late} and adopt ResNet-18 and ResNet-34 for CIFAR-10 and CIFAR-100, respectively. For all noisy conditions on CIFAR-10 \& 100, we leverage an SGD optimizer with the momentum $0.9$ and the weight-decay $1\times10^{-3}$ to train our network. The total training epoch is set as 200. The initial learning rate is 0.02 and decayed with the factor 10 at the 100-th and 150-th epoch. For real-world noise, we adopt a ResNet-34 for CIFAR-N and a pre-trained ResNet-50 for Clothing-1M and Food101N, and Inception-ResNet-V2 \cite{szegedy2017inception} for WebVision. Other training set-ups like the number of training epochs, the learning rate and its adjustment, the optimizer, and so on keep the same as \cite{li2024sure}.

\noindent\textbf{Hyper-parameters}. Our framework contains two main hyper-parameters, \textit{i.e.}, the number of expert $m$ in the network architecture, and the slope parameter $\beta$ in the mapping function. We keep $\beta=2$ for all experiments and set $m=4$ for CIFAR and $m=2$ for three open-world datasets.

\subsection{Results on Synthetic Noisy Datasets}
We conduct comparison experiments with two synthetic noise types on CIFAR-10 \& 100, where we manually construct different noisy types on these two datasets. Since our framework is agnostic to selection criteria, we test our proposal on three types of selection criteria.  Experimental results are shown in Table \ref{tab:results_onCIFAR}. 

On both CIFAR-10 and CIFAR-100, our method ITEM consistently achieves state-of-the-art performance in all noisy settings. Compared with other section-based methods such as Co-teaching and SFT, ITEM obtains obvious performance improvements while adopting their selection criteria. To be specific, $\dagger$ITEM averagely improves the test accuracy of Co-teaching by 9.68\% on CIFAR-10 and 22.06\% on CIFAR-100. 
Compared with SFT \cite{wei2022self}, a representative method in fluctuation-based selection, ITEM also achieves remarkable improvements.

\begin{table}[t]
\centering
\caption{\small Test accuracy (\%) comparison of previous methods on three real-world noisy benchmarks. Top-1 and Top-5 test accuracy are reported for the dataset WebVision.}
\vspace{-3mm}
\resizebox{0.46\textwidth}{!}{
\setlength{\tabcolsep}{1.3mm}{
\begin{tabular}{l|ccccc}
\toprule[0.85pt]
\multicolumn{1}{c|}{\multirow{2}{*}{Methods}}         & \multirow{2}{*}{Food101N}     & \multirow{2}{*}{Clothing1M}         & \multicolumn{2}{c}{WebVision}  \\ 
                &               &                   &  Top-1   & Top-5  \\  \midrule
Co-teaching \cite{han2018co}                & 83.73         & 67.94             & 63.58 & 85.20 \\
JoCoR \cite{wei2020combating}               & 84.04         & 69.06             & 63.33 & 85.06 \\
CDR \cite{xia2020robust}                    & 86.36         & 66.59             & 62.84 & 84.11  \\
ELR+ \cite{liu2020early}                    & 85.77         & 74.81             & 77.78 & 91.68  \\
DivideMix \cite{li2020dividemix}            & 86.73         & 74.76             & 77.32 & 91.64  \\
SFT+  \cite{wei2022self}                    & -             & 75.00             & 77.27 & 91.50  \\
CoDis* \cite{xia2023combating}              & 86.88         & 74.92             & 77.51 & 91.95   \\
SURE \cite{li2024sure}                      & 88.00         & \textbf{75.10}             & 78.94 & 92.34  \\ \midrule
$\ddagger$ ITEM*                            & \textbf{88.14}        & 75.08             & \textbf{80.20} & \textbf{93.07} \\
\bottomrule[0.85pt]  
\end{tabular}
}}
\label{tab:realworld}
\end{table}

\subsection{Results on Real-World Noisy Datasets}
We test ITEM's performance on two human-made noisy datasets and three website noisy datasets. Considering that semi-supervised learning can utilize the sample discarded by the sample selection framework to regularize the model's learning \cite{li2020dividemix}, we integrate ITEM with the Pseudo-labeling technique \cite{lee2013pseudo} to further improve the performance.

1) \emph{Results on CIFAR10N \& 100N}. The results are shown in Table \ref{tab:cifar10N}. First, our method achieves remarkable performance on both two settings (\emph{w/} or \emph{w/o} SSL). When training without SSL, ITEM outperforms previous methods by a large margin. Compared with Late Stopping, a newly proposed method, ITEM achieves 5.88\% improvements on \emph{worst}-labels. When training with SSL, ITEM is superior to the previous SOTA method, DivideMix. Even training without SSL, ITEM surprisingly gains greater performance than those methods that leverage SSL approaches.

2) \emph{Results on Food101N, Clothing1M, and WebVision}. The results are shown in Table \ref{tab:realworld}. On Food101N, our proposal achieved the best performance, outperforming SURE (the method with the second-best performance) \cite{li2024sure} by 0.14\%. On Clothing1M, the performance of our proposal is not the best but considerably competitive. Compared with the best performance, the 75.1\% test accuracy, ITEM obtained a closed score, \emph{i.e.}, 75.08\%. On WebVision, we keep the convention from the work \cite{xia2023combating} that tests the performance on the validation set of WebVision. ITEM significantly improved Top-1 accuracy by nearly 1.3\%. 

The result verifies that our method effectively enhances model generalization in large-scale real-world scenarios.

\begin{figure*}[t]
    \centering
    \includegraphics[width=0.98\linewidth]{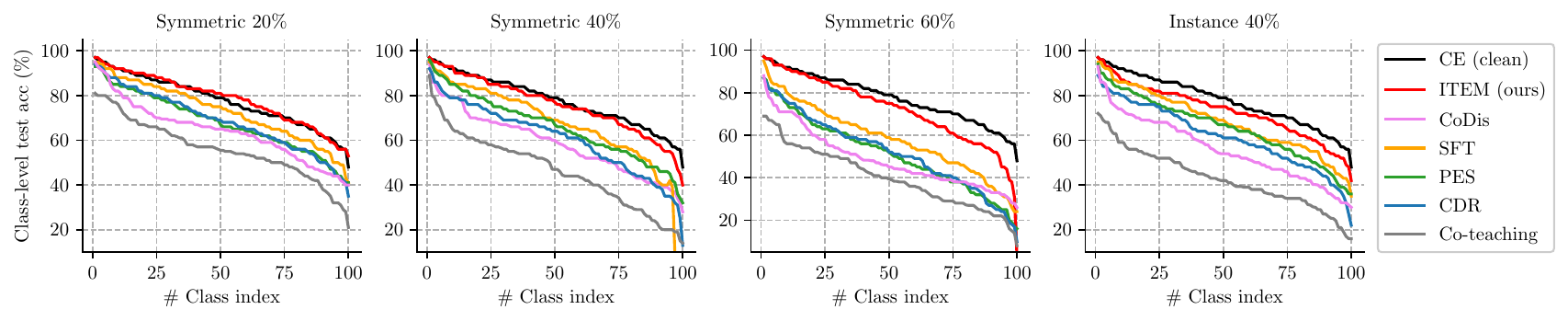}
    \vspace{-2mm}
    \caption{\small \textbf{Visualization of debias learning} in a class-level. We selected a ResNet-18 as the backbone and compared class-level prediction results of various methods on CIFAR-100 with four noise types. ``CE (clean)" denotes training the model on the completely clean set (50k samples in total). The index of classes is sorted according to the class-level accuracy.
    }
    \vspace{-2mm}
    \label{fig:debias_curve}
\end{figure*}

\begin{figure}[t]
\centering
\subfloat[\footnotesize Small-loss selection \cite{han2018co}]{
\begin{minipage}[b]{0.98\linewidth}
\centering
\includegraphics[width=0.95\linewidth]{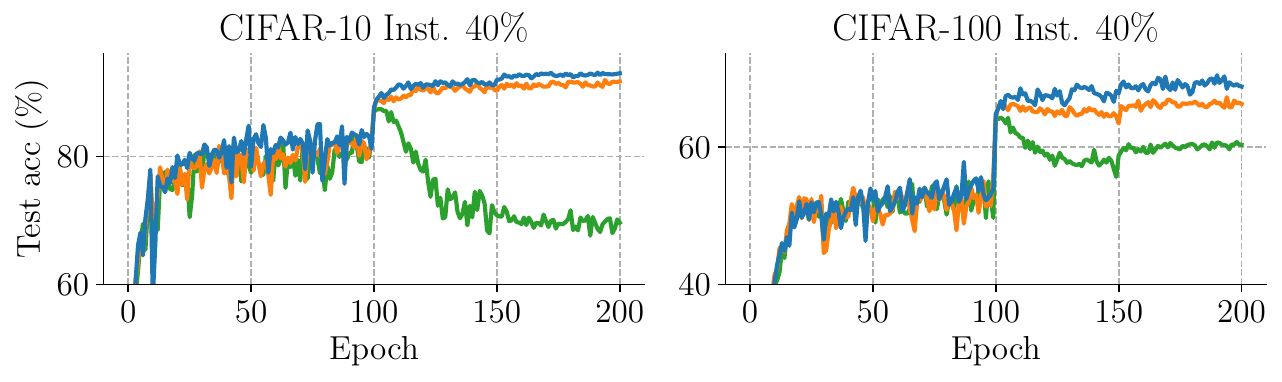}
\end{minipage}}
\vspace{2mm}
\\
\subfloat[\footnotesize Fluctuation-based selection \cite{wei2022self}]{
\begin{minipage}[b]{0.98\linewidth}
\centering
\includegraphics[width=0.95\linewidth]{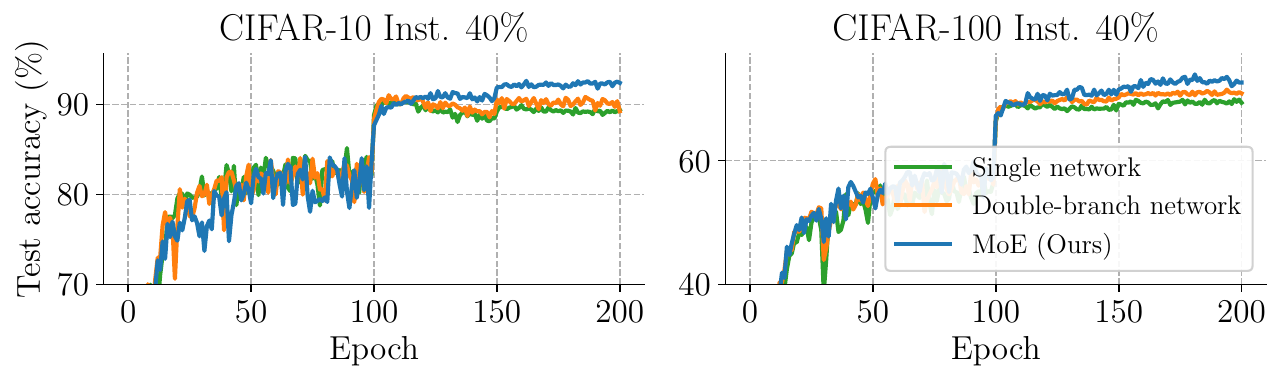}
\end{minipage}}
\vspace{-1mm}
\caption{\small \textbf{Robustness comparison} of different networks under two selection criteria. Our proposed MoE structure significantly mitigates the adverse effect of noisy labels.}
\label{fig:network}
\end{figure}

\subsection{More Analyses}

\noindent\textbf{Debias learning for LNL.} 
We visualize the test accuracy in each class to demonstrate that ITEM achieves relatively balanced performance on varying categories while fulfilling greater generalization. We plot comparison results in Figure \ref{fig:debias_curve}. First, ITEM achieves almost unbiased prediction results compared to \emph{CE (clean)}, which is trained on 50k clean samples. Second, compared with existing methods, ITEM takes tail classes better into account since two weighted samplers. Therefore, ITEM performs better in these categories. 
Besides, under 20\% symmetric label noise, ITEM achieved better performance than the model trained on an absolutely clean set (see the red line \textit{vs.} the black line).
These results demonstrate that our approach has significant potential to address the implicit bias inherent in sample selection frameworks for noisy label learning.



\noindent\textbf{Robustness of MoE structure.} Compared with existing robust network, our proposal MoE exhibits better robustness. To clearly show this merit, we evaluate two selection criteria and plot the training curve under three networks in Figure \ref{fig:network}. 
For the small-loss criterion, which easily incurs accumulation errors (see the green line (a)), the performance improvement with the MoE structure is significant. In contrast to the double-branch structure, which experiences slight performance degradation when the learning rate decreases at the 100th epoch (see the orange line in (b)), the learning curve with the MoE structure is more robust and stable.
Second, in terms of fluctuation-based selection, the MoE demonstrates greater robustness, more stable training curves, and better generalization performance in noisy settings. In addition, its plug-in plug-out capability further verifies the versatility of the MoE structure.

\begin{table}[t]
\centering
\caption{\small \textbf{Ablation studies} of each component in ITEM with varying noise conditions.}
\vspace{-2mm}
\resizebox{0.48\textwidth}{!}{
\setlength{\tabcolsep}{2mm}{
\begin{tabular}{l|cccc}
\toprule[0.9pt]
\multirow{2}{*}{Methods}  & \multicolumn{1}{c}{CIFAR-10}  & \multicolumn{1}{c}{CIFAR-100}   & CIFAR10N  & CIFAR100N\\ 
    & {Sym 40\%} & {Sym 40\%}  & {Worst}  & {Fine}\\ \midrule
 \emph{w/o} MoE Network & 90.75  & 73.01  & 89.72 & 68.19\\ 
 \emph{w/o} Mixed Sampling   & 88.77  & 71.09  & 87.66  & 68.24\\  
 \emph{w/o} MixUp  & 91.19 & 72.50  & 89.90 & 69.10\\
 Ours & \textbf{92.81}  & \textbf{74.90}  & \textbf{91.15}  & \textbf{69.72} \\ 
\bottomrule[0.9pt]  
\end{tabular}
}
}
\label{tab:ablation}
\end{table}

\subsection{Ablation Studies}

\noindent\textbf{Effect of each component.} Our framework mainly contains two modules: the robust network architecture and the mixed data sampling strategy. We conduct ablation studies on CIFAR benchmarks to evaluate the effectiveness of each component from the following two perspectives. 

\textit{1) Quantitative analysis.} We split the main component in ITEM into three parts: robust MoE, Mixed sampling, and the MixUp strategy. By solely removing them from ITEM, we reported the performance of the model in Table \ref{tab:ablation}. Compared with typical ResNet architecture, our expert-based network indeed makes an obvious contribution to mitigating the noise. The average improvement among five settings roughly reaches 2\%. Second, the mixed sampling strategy also promotes the performance of our network. Compared with training on randomly sampled batches, training on two weighted sampled batches gains greater generalization. 

\begin{figure}[t]
\flushleft
\vspace{1mm}
\subfloat[\footnotesize $acc=88.94\%$]{
\begin{minipage}[b]{0.322\linewidth}
\centering
\includegraphics[width=0.98\linewidth]{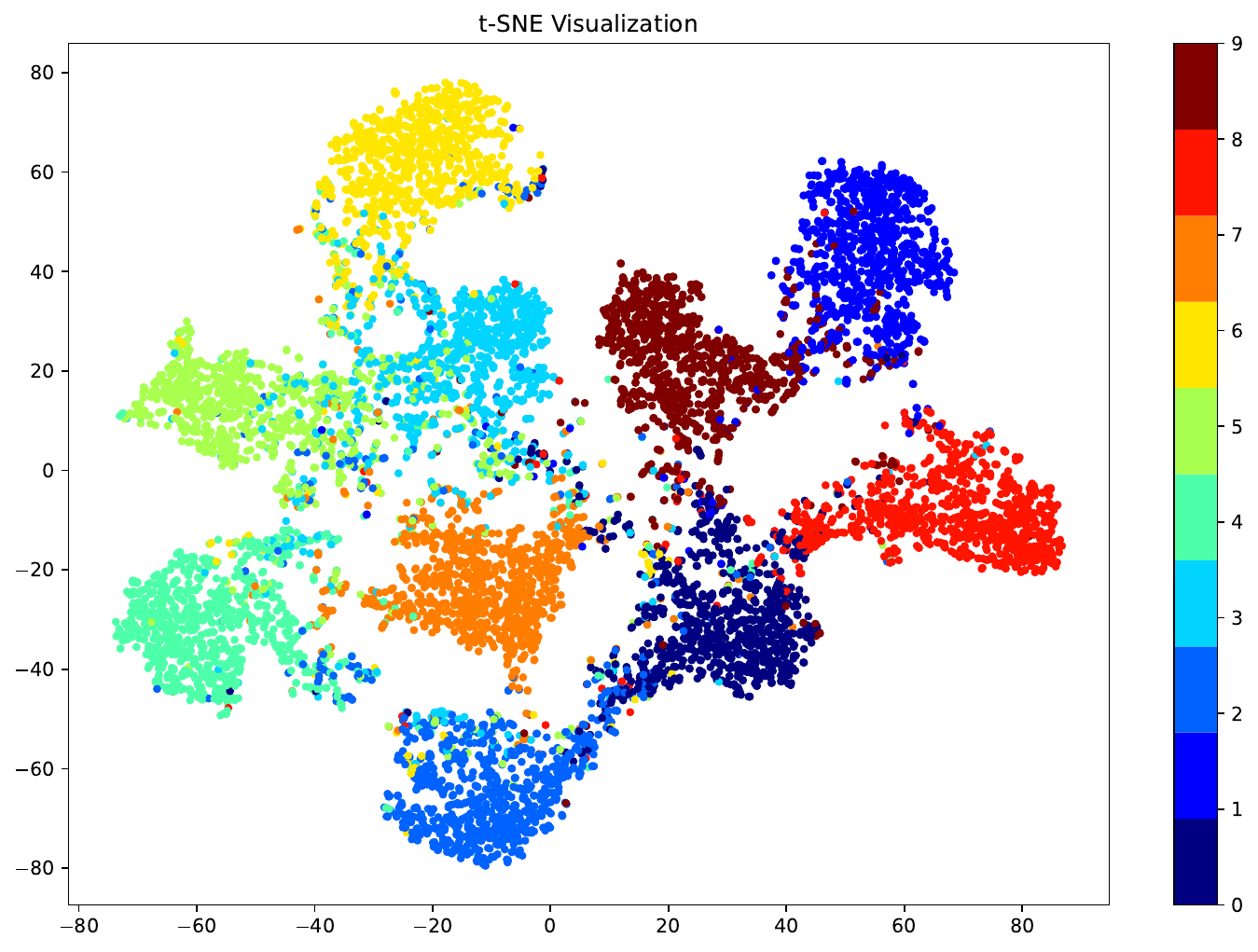}
\end{minipage}}
\hfil
\subfloat[\footnotesize $acc=91.29\%$]{
\begin{minipage}[b]{0.322\linewidth}
\centering
\includegraphics[width=0.98\linewidth]{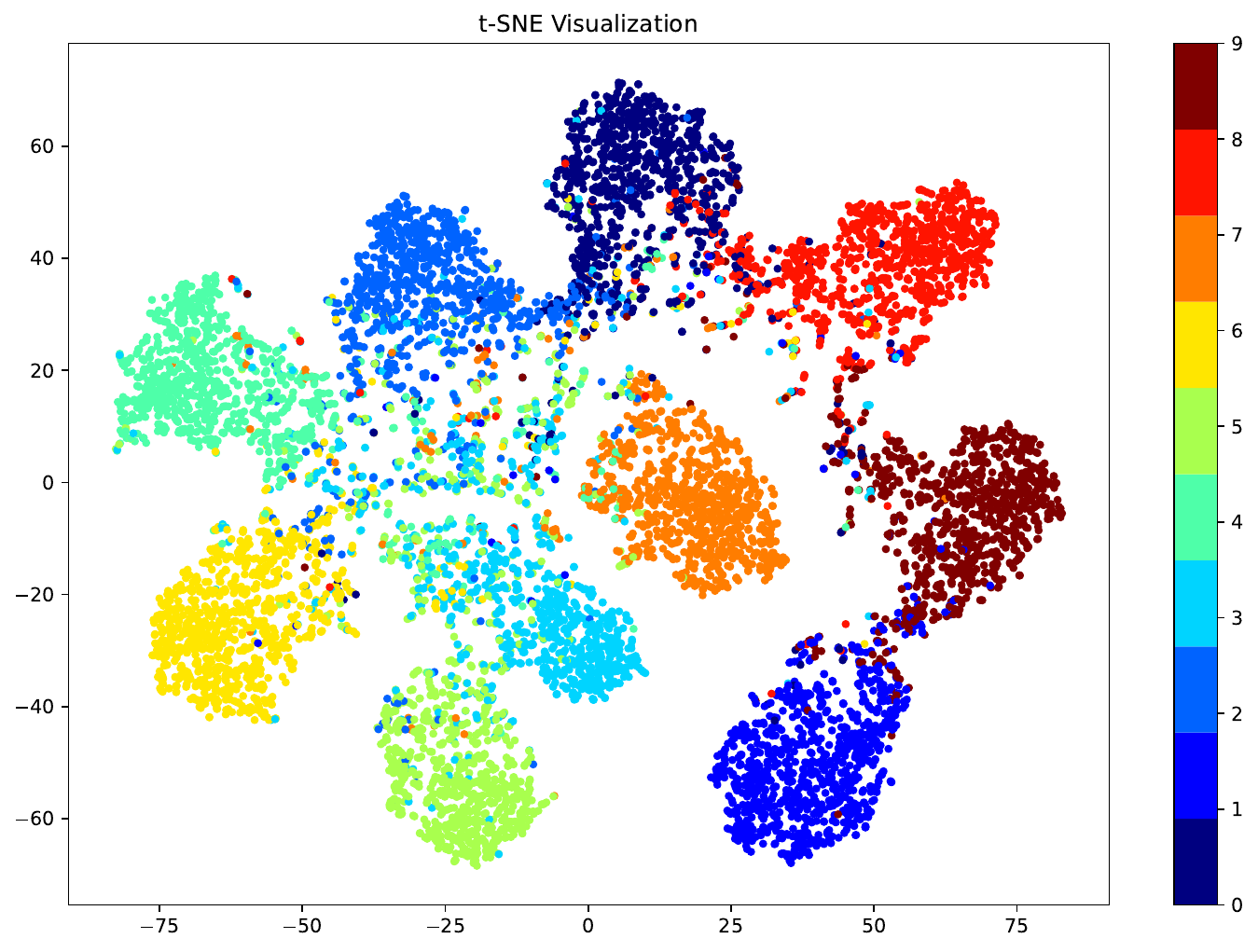}
\end{minipage}}
\hfil
\subfloat[\footnotesize $acc=92.97\%$]{
\begin{minipage}[b]{0.322\linewidth}
\centering
\includegraphics[width=0.98\linewidth]{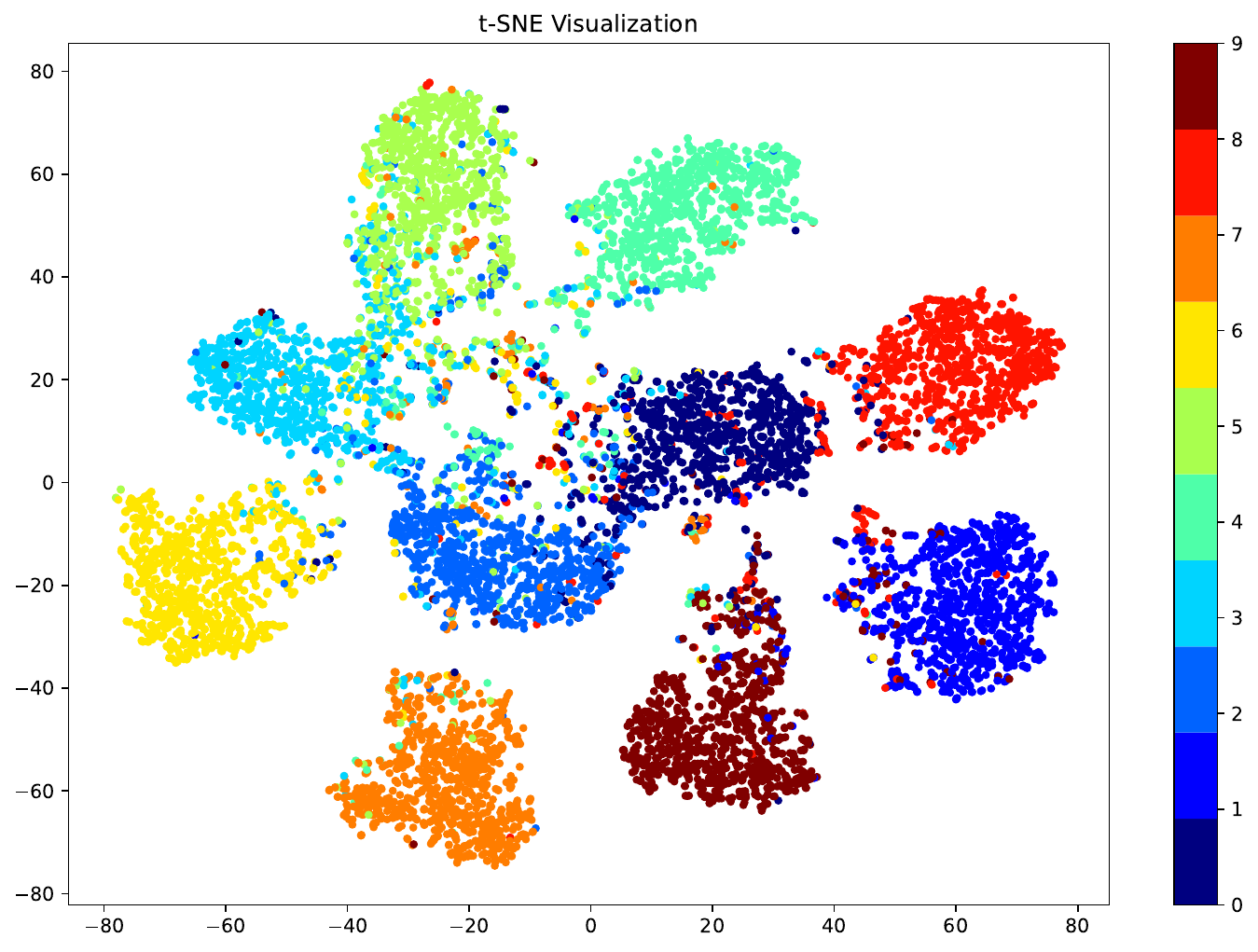}
\end{minipage}}
\caption{\small Visualization of feature representations on the CIFAR-10 test set by T-SNE \cite{van2008visualizing} with \textbf{different sampling strategies}. (\textbf{a}) Training on randomly sampled batch $B$, (\textbf{b}) Training on $B_{\tilde v}$, and (\textbf{c}) Training on $\texttt{MixUp}(B_{\bm{v}}, B_{{\bm{\tilde v}}})$.}
\label{fig:TSNE}
\end{figure}

\textit{2) Representation visualization}. Considering that training with different sampling leads the model to different representation performance, we compare the learned representation space of the test set of CIFAR-10 and show the results in Figure \ref{fig:TSNE}. We can see that the learned representation of ten classes on the normal sampling $B$ is incompact, while the decision boundary is not well generalized. By contrast, if the proposed reversed tail-focused batch $B_{\tilde v}$ is adopted, each cluster is more compact where the distance between each cluster is larger, which verifies the effectiveness of our proposal sampling strategy in mitigating data bias underlying the selection phase. Eventually, by leveraging the Mixup operation, a feature extractor with better representation performance is obtained, as shown in the last plot.


\begin{figure}[t]
\centering
\subfloat[\footnotesize Number of experts $m$]{
\begin{minipage}[b]{0.48\linewidth}
\centering
\includegraphics[width=0.98\linewidth]{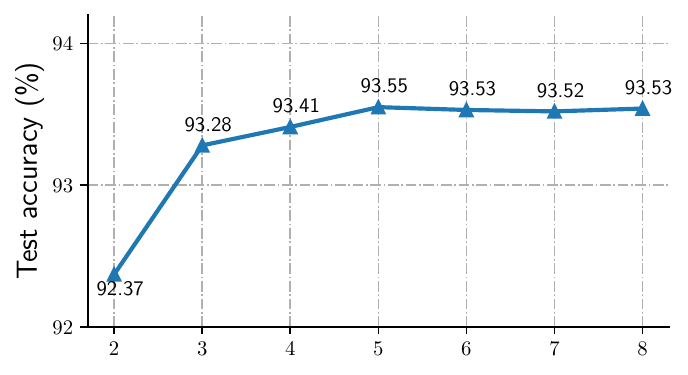}
\end{minipage}}
\subfloat[\footnotesize The slope parameter $\beta$]{
\begin{minipage}[b]{0.48\linewidth}
\centering
\includegraphics[width=0.98\linewidth]{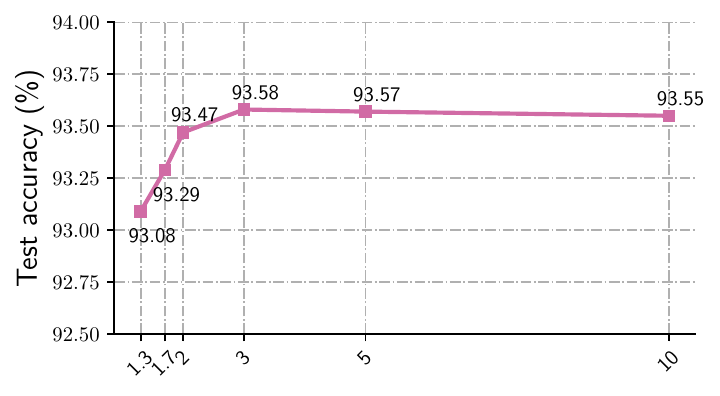}
\end{minipage}}
\vspace{-2mm}
\caption{\small \textbf{Sensitivity analyses} of two hyper-parameters in ITEM. A ResNet-18 is trained on CIFAR-10 with 40\% sym. noisy labels.}
\label{fig:sensitivity}
\vspace{-2mm}
\end{figure}

\noindent\textbf{Sensitivity analyses.} ITEM has two main hyper-parameters, $m$ and $\beta$. We conduct ablation studies to select optimal values for experiments. The result is shown in Figure \ref{fig:sensitivity}. First, the number of expert layers $m$ is the most important parameter in our framework, which decides the overall network structure and computation costs. From the right figure, we can observe that the performance of ITEM gradually increases with the value of $m$ increases. When arriving at a peak, i.e., 93.55\% top-1 accuracy at $m=5$, the performance will not increase if the value of $m$ becomes larger. Therefore, we recommend a larger value of $m$ given an unknown dataset. Second, the value of $\beta$ has slightly influenced the performance of our method, where we also prefer a larger value of $\beta$.

%% file: 2_related_new.tex
There are three main research lines in sample selection against the incorrect information from noisy labels.

\emph{1) Robust selection criteria}. Previous sample selection methods normally exploit the memorization effect of DNNs, i.e., DNNs first memorize training data with clean labels and then those with noisy labels \cite{han2018co,xia2020robust}. The small-loss criterion \cite{han2018co} is a typical method stemming from the memorization effect, which splits the noisy training set via a loss threshold. The samples with small losses are regarded as clean samples. Some improved criteria \cite{li2020dividemix,arazo2019unsupervised} based on the small-loss criterion are proposed in later works. 
In addition to the family of small loss criterion, some methods motivated by prediction fluctuation are designed. These methods  \cite{xia2021sample,zhou2020robust} observe that the model tends to give inconsistent prediction results for noisy samples and thus filter noisy labels by identifying high-frequency fluctuation samples.

\emph{2) Robust network architecture}. Easily works based on constructing the noise transition matrix always resort to an additional adaptation layer at the top of the softmax layer \cite{jindal2016learning} or design a new dedicated architecture \cite{han2018masking,yao2018deep}. Recently, a family of double-branch networks \cite{han2018co,yu2019does,wei2020combating,li2020dividemix} has been proposed, which integrates two networks with the same architecture and selects clean samples for another network. Besides, better performance would be achieved when ensembling prediction results from two networks \cite{han2018co}.

\emph{3) Robust training procedures}. Overall, current methods in sample selection always resort to a self-training manner, i.e., iteratively conducting clean label selection and model retraining. However, the incorrect selection would degrade the subsequent model learning. 
Recently, some robust training manners were proposed. Concretely, \cite{zhou2020robust} designed a framework based on curriculum learning \cite{bengio2009curriculum}, which starts with learning from clean data and then gradually moves to learn noisy-labeled data with pseudo labels. 

\noindent\textbf{Relations to us.} The advantages of our proposal compared with previous methods can be divided into the following three parts, which are summarized in Table \ref{tab:related_work} in Appendix.
\begin{itemize}
\setlength\itemsep{0mm}
    \item \textit{Flexibility of the network structure}. To our knowledge, we are the first to apply a mixture-of-experts (MoE) structure to LNL. Despite the core idea of MoE and double-branch structure being similar (\textit{i.e.},ensembling prediction results and the selection result over different views), MoE exhibits more flexibility. Expanding the number of experts is simple and only increases slight parameter counts (0.01 million per expert). By contrast, 
    the double-branch structure is hardly extended to a three- or four-branch structure since the number of learnable parameters. 
    
    \item \textit{Expansibility of the framework}. Compared with previous selection-based methods, our proposal is agnostic to the selection criteria, which can be easily integrated with different selection criteria and improves their performance.
    \item \textit{Debias learning}. Previous mainly focus on the accumulation error, a training bias. In this paper, we empirically show the existence of data bias (\textit{i.e.}, the imbalanced dataset selected by current selection criteria) and design an expert-based sampling strategy, which mitigates this bias by adjusting selection probabilities and increasing the visibility of underrepresented classes.
\end{itemize}

%% file: 6_conclusion.tex
In this paper, we disclosed the data bias, an implicit bias underlying the sample selection strategy. 
To solve the training and data bias simultaneously, we proposed ITEM that introduces a noise-robust multi-experts network and a mixed sampling strategy.
First, our structure integrates the classifier with multiple experts and leverages expert layers to conduct selection independently. Compared with the prevailing double-branch network, it exhibits great potential in mitigating the training bias, i.e., the accumulated error.
Second, our mixed sampling strategy yields class-aware weights and further conducts weighted sampling. 
The effectiveness of ITEM in real applications is verified on diverse noise types and both synthetic and real-world datasets.

%% file: appendix.tex
\onecolumn  
\appendix   
\vspace{3mm}


\setcounter{section}{0}
\renewcommand{\thesection}{\Alph{section}}

\section{Visualization of Data Bias}
In this section, we conducted extensive experiments that discovered the existence of the data bias, \textit{e.g.}, the imbalanced subset is selected by current selection-based frameworks. Despite Figure \ref{fig:imbalanced_intro} in the section on Introduction, we show more results in the following figure (See Figure \ref{fig:imbalanced_intro2}).
We can observe that \textit{data bias widely exists in selection-based frameworks no matter the different noise types and different datasets.}

\begin{figure*}[h]
\centering
\subfloat[]{
\hspace{-6mm}
\begin{minipage}[b]{0.33\linewidth}
\centering
  \includegraphics[width=0.98\linewidth]{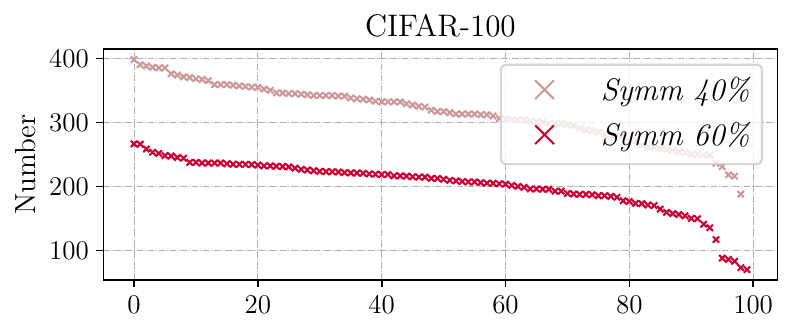}\\ \includegraphics[width=0.98\linewidth]{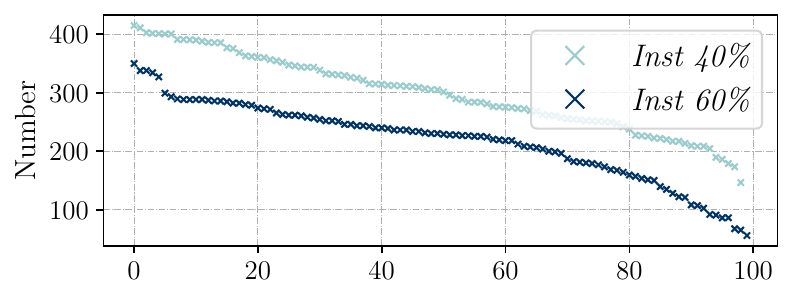}  \\
\includegraphics[width=0.98\linewidth]{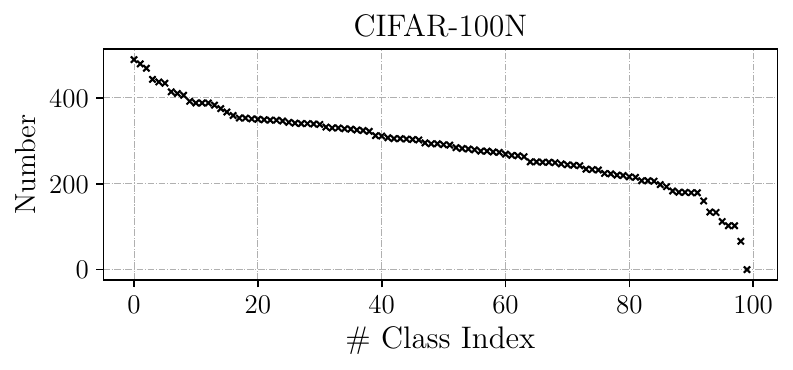}  
\end{minipage}
}
\subfloat[]{
\hspace{-2mm}
\begin{minipage}[b]{0.33\linewidth}
\centering
 \includegraphics[width=0.98\linewidth]{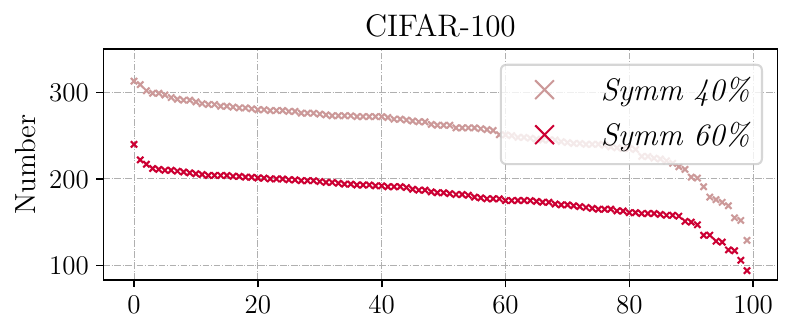}\\
 \includegraphics[width=0.98\linewidth]{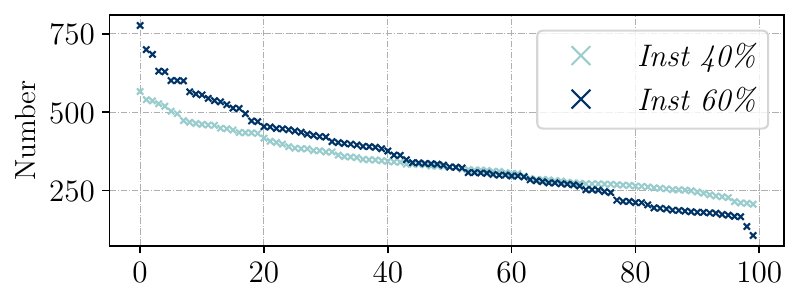}\\
  \includegraphics[width=0.98\linewidth]{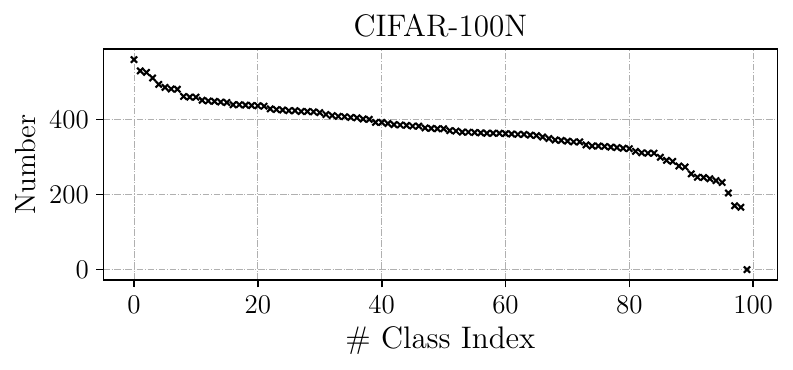}
\end{minipage}
}
\subfloat[]{
\hspace{-2mm}
\begin{minipage}[b]{0.33\linewidth}
\centering
 \includegraphics[width=0.98\linewidth]{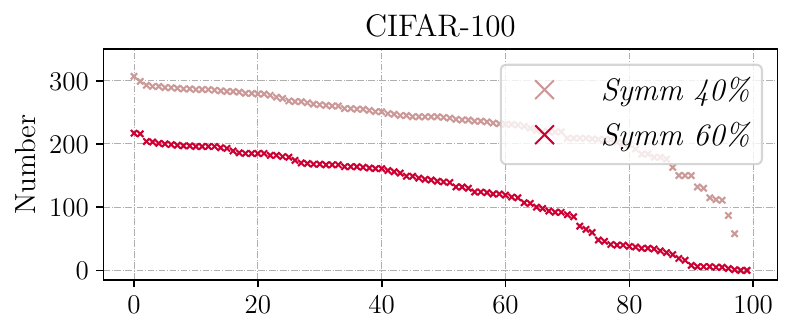}\\
 \includegraphics[width=0.98\linewidth]{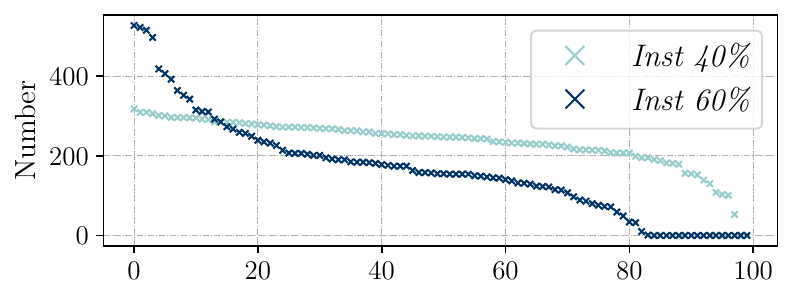}\\
  \includegraphics[width=0.98\linewidth]{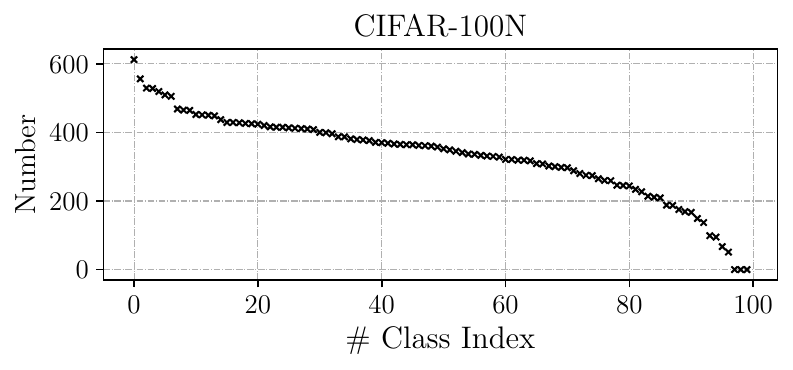}
\end{minipage}
}
\caption{\textbf{Extended results} of Figure \ref{fig:imbalanced_intro}. (a) Small-loss with threshold \cite{han2018co}, (b) Small-loss with GMM \cite{li2020dividemix}, and (c) Fluctuation-based noise filtering \cite{wei2022self}.}
\label{fig:imbalanced_intro2}
\end{figure*}

\begin{table*}[h]
\centering
\caption{We compare ITEM with various selection-based methods over four perspectives.}
\label{tab:related_work}
\scalebox{.92}{
\setlength{\tabcolsep}{1mm}{
\begin{tabular}{c|cccc}
\toprule[1pt]
Methods                      & Network architecture   & Selection-criterion specific    & Training debias  & Data debias\\ \midrule
Co-teaching \cite{han2018co}, CoDis \cite{xia2023combating}          & Double-branch   & Small-loss criterion            &  \ding{55}       &  \ding{55}\\
JoCoR \cite{wei2020combating}, DivideMix \cite{li2020dividemix}, ProMix \cite{wang2022promix}                   & Double-branch   & Small-loss criterion            &  \ding{51}       &  \ding{55}\\
SFT \cite{wei2022self}, Late Stopping \cite{yuan2023late}                  & Single-branch   & Fluctuation-based selection     &  \ding{51}       &  \ding{55} \\         \midrule
ITEM {\small (ours)}                & Mixture-of-Experts  & Not specific                    &  \ding{51}       & \ding{51} \\
\bottomrule[1pt]  
\end{tabular}}
}
\end{table*}




\section{Algorithm of ITEM}\label{appendixsec:ssl}
We provided the pseudocode for ITEM in Algorithm \ref{alg:algorithm}.

\begin{algorithm}[h]
   \caption{ITEM}
   \label{alg:algorithm}
{\bfseries Input:} Noisy training set ${D}_N$, training epoch $T$, warmup iterations $T_{w}$, a selection criterion $\textsf{Criterion}(\cdot)$, a classifier network with $m$ experts, and the mapping function $\mathcal{S^\beta(\cdot)}$ with a hyper-parameter $\beta$. 
\begin{algorithmic}[1]
    \STATE {\bfseries Initialize} our network with one classifier layer and $m$ expert layers $\mathbbm{F} = \{f, g^1,...,g^m\}$.   \\
    \FOR{$t$ = 1, ..., $T_w$} 
        \STATE Randomly sample $f^c$ from $\mathbbm{F}$, and WarmUp $f^c$ on $D_N$. \\
    \ENDFOR
    \FOR{${t}$ = 1, ..., $T$}
        \STATE Select the clean set ${D}_{\mathrm{clean}}$ from $D_N$ based on $\textsf{Criterion}(\mathbbm{F})$.\\
        \STATE Calculate weighted vector $\bm{v} = [{w}_1, ..., {w}_K]$ on ${ D}_{\mathrm{clean}}$. \\ 
        \STATE Calculate reversed weighted vector $\bm{\tilde v} = [{\tilde w}_1, ..., {\tilde w}_K]$ with Equation (\ref{eq:mapping}). \\ 
        \WHILE{$i$ = 1, ..., iterations}
            \STATE Sample a mini batch $B_{{\bm{v}}} = \{(\bm{x}_i, {\tilde y}_i)\}_{i=1}^b$ from ${D}_{\mathrm{clean}}$ according to $\bm{v}$.  \\
            \STATE Sample a mini-batch $B_{{\bm{\tilde v}}} = \{(\bm{x'}_i, {\tilde y'}_i)\}_{i=1}^b$ from ${D}_{\mathrm{clean}}$ according to $\bm{\tilde v}$. \\
            \STATE Sample $f^c$ from $\mathbbm{F}$. \\
            \STATE Update the network parameters by minimizing the loss in Equation (\ref{eq:trainLoss}).
        \ENDWHILE
   \ENDFOR
\end{algorithmic}
\end{algorithm}

\noindent \textbf{ITEM with semi-supervised learning.} Considering the large scale of unreliable samples discarded, especially in considerable noise ratios, leveraging these samples through semi-supervised learning (SSL) to regularize the model is necessary as well as effective. \textit{Thanks to the flexible training framework, any SSL method can be integrated into our proposal ITEM.} 

Here, we provide a novel version of ITEM with stronger performance. The algorithm is shown in Algorithm \ref{alg:algorithm2}. Besides, a strong augmentation \texttt{CTAugment} is used, which refers to \cite{sohn2020fixmatch}.
Compared with its original version, the new training framework only adds two operations (shown in Lines 12 and 13). Thus, the code implementation is relatively convenient.

\begin{algorithm}[t]
   \caption{ITEM with a semi-supervised framework}
   \label{alg:algorithm2}
{\bfseries Input:} The training set $D_N$, training epoch $T$, warmup iterations $T_{w}$, a selection criterion $\textsf{Criterion}(\cdot)$, a classifier network with $m$ experts, the mapping function $\mathcal{S^\beta(\cdot)}$ with a hyperparameter $\beta$.  
\begin{algorithmic}[1]
    \STATE {\bfseries Initialize} our network with one classifier layer and $m$ expert layers $\mathbbm{F} = \{f, g^1,...,g^m\}$.   \\
    \FOR{$t$ = 1, ..., $T_w$} 
        \STATE Randomly sample $f^c$ from $\mathbbm{F}$, and WarmUp $f^c$ on $D_N$. \\
    \ENDFOR
    \FOR{${t}$ = 1, ..., $T$}
        \STATE Split $D_N$ into the clean set ${D}_{\rm clean}$ and the unreliable set ${\tilde D}_{\rm noise}$ based on $\textsf{Criterion}(\mathbbm{F})$, and discard labels in ${\tilde D}_{\rm noise}$.\\
        \STATE Calculate weighted vector $\bm{v} = [{w}_1, ..., {w}_K]$ on ${\tilde D}_{ clean}$. \\ 
        \STATE Calculate reversed weighted vector $\bm{\tilde v} = [{\tilde w}_1, ..., {\tilde w}_K]$ with Eq. (\ref{eq:mapping}). \\ 
        \WHILE{$i$ = 1, ..., iterations}
            \STATE Randomly sample a batch $B_{{\bm{v}}} = \{(\bm{x}_i, {\tilde y}_i)\}_{i=1}^b$ from ${D}_{\rm clean}$ according to $\bm{v}$.  \\
            \STATE Randomly sample a batch $B_{{\bm{\tilde v}}} = \{(\bm{x}'_i, {\tilde y'}_i)\}_{i=1}^b$ from ${D}_{\rm clean}$ according to $\bm{\tilde v}$. \\
            \STATE Draw a mini-batch  ${\hat B} = \{(\bm{x}''_i)\}_{i=1}^{2b}$ from ${D}_{\rm noise}$ without labels. \\
            \STATE Generate pseudo-labels for each sample in ${\hat B}$, then have ${\hat B} = \{(\bm{x}''_i, {\tilde y''}_i)\}_{i=1}^{2b}$, where ${\tilde y''}_i = \mathbbm{E}_{f \sim \mathbbm{F}}[f(\bm{x}''_i)]$. \\
            \STATE Randomly sample $f^c$ from $\mathbbm{F}$. \\
            \STATE Update network parameters via minimizing $\mathcal{L} \big{(}f^c \big{(}\texttt{MixUp}(B_{{\bm{v}}} \text{ and } B_{{\bm{\tilde v}}}, {\hat B})  \big{)}$.
        \ENDWHILE
   \ENDFOR
\end{algorithmic}
\end{algorithm}

\section{Experimental Settings}\label{appsec:experiments}
In this section, we reported the statistical data of seven datasets utilized in this paper and the whole experimental setup in Table \ref{tab:train_setup}. 

\subsection{Dataset}

\textbf{CIFAR-10 and CIFAR-100} \cite{krizhevsky2009learning} dataset consists of 60,000 images of 10 and 100 categories. Each image is with a resolution of 32 $\times$ 32. 

\textbf{CIFAR10N and CIFAR100N} \cite{wei2021learning} also consists of 60,000 images. Meanwhile, the noise labels in these two are human-made with varying noise conditions, including \emph{worst} and \emph{random 1\&2\&3} for CIFAR10N and \emph{fine noise} for CIFAR100N. 

\textbf{Clothing1M}~\cite{xiao2015learning} is a large-scale dataset that is collected from real-world online shopping websites. It contains 1 million images of 14 categories whose labels are generated based on tags extracted from the
surrounding texts and keywords, causing huge label noise. The estimated percentage of corrupted 
labels is around 38.46\%. A portion of clean data is also included in Clothing1M, which has been divided into the training set (903k images), validation set (14k images), and test set (10k images). We resize all images to $256 \times 256$ as in~\citep{li2020dividemix} and then random crop to 224 $\times$ 224.

\textbf{Food-101N} \cite{lee2018cleannet} is constructed based on the taxonomy of $101$ categories in Food-101~\citep{bossard2014food}. It consists of 310k images collected from Google, Bing, Yelp, and TripAdvisor. The noise ratio for labels is around 20$\%$. Following the testing protocol in~\citep{lee2018cleannet}, we learn the model on the training set of 55k images and evaluate it on the testing set of the original Food-101. 

\textbf{WebVision} \cite{li2017webvision}. Following \cite{liu2020early}, we use a mini version of WebVision where only the top 50 classes are utilized. This mini WebVision dataset contains approximately 66 thousand images and the corresponding noise ratio is roughly 20\%.

\subsection{Other Baselines}
On two human-annonated datasets CIFAR10N and CIFAR100N, we compare ITEM with Co-teaching+ \cite{yu2019does}, Peer Loss \cite{liu2020peer}, CAL \cite{zhu2021second}, DivideMix \cite{li2020dividemix}, ELR+ \cite{liu2020early}, CORES* \cite{cheng2020learning}, and DPC \cite{zong2024dirichlet}. The results are collected from the literature \cite{wei2021learning}. On three real-world noisy datasets, we compared ITEM with SURE \cite{li2024sure}, Co-teaching \cite{han2018co}, MentorNet \cite{jiang2018mentornet}, JoCoR \cite{wei2020combating}, CDR \cite{xia2020robust}, ELR \cite{liu2020early}, DivideMix \cite{li2020dividemix}, SFT \cite{wei2022self}, CoDis \cite{xia2023combating}, SURE \cite{li2024sure}. 




\begin{table*}[t]
\centering
\small
\caption{Experimental settings about training procedure of \textit{ITEM}.} 
\resizebox{1\textwidth}{!}{
\setlength{\tabcolsep}{2mm}{
\begin{tabular}{l|c|c|c|c|c|c|c}
\toprule[0.9pt]
Datasets        & CIFAR-10       & CIFAR-100    & CIFAR-10N  & CIFAR-100N & Clothing-1M  & Food-101N & WebVision   \\  \midrule
Class number    & 10           & 100          & 10        & 100         & 14        & 101       & 50        \\ 
Training size   & 50,000       & 50,000       & 50,000    & 50,000      & 1,000,000 & 310,009   & 66,000    \\ 
Testing size    & 10,000       & 10,000       & 10,000    & 10,000      & 10,000    & 25,250    & 2,500      \\ \midrule
\multicolumn{5}{l}{\textbf{Training procedure}} \\ \midrule
Network           & ResNet-18    & \multicolumn{3}{c|}{ResNet-34}   & \multicolumn{2}{c|}{ResNet-50}       & InceptionResNetV2 \\ \cline{2-8}
Batch size      & \multicolumn{4}{c|}{64}               & 100       & \multicolumn{2}{c}{32}   \\ \cline{2-8}
Epoch           &  \multicolumn{4}{c|}{200}             & 10        & 100   & 30  \\ \cline{2-8}
Warmup epoch    & 10           & 30           & 10     & 30       & 1     & 5   & 3\\ \cline{2-8}
Learning rate (LR)  & \multicolumn{4}{c|}{0.02}             & \multicolumn{2}{c|}{0.001}    & 0.02  \\ \cline{2-8}
Weight decay    & \multicolumn{4}{c|}{1e-3}             & 5e-4   \\ \cline{2-8}
LR scheduler    &\multicolumn{4}{c|}{divide 10 at [100,150]th epoch}  & divide 10 at 5th epoch & divide 10 at 50th epoch & divide 10 at 20th epoch\\ \cline{2-8}
Optimizer       & \multicolumn{7}{c}{SGD} \\ \cline{2-8}
Momentum        & \multicolumn{7}{c}{0.9} \\  \midrule
\multicolumn{5}{l}{\textbf{Hyperparameters}} \\ \midrule
experts number $m$     & \multicolumn{4}{c|}{$m=4$}  & $m=2$ & \multicolumn{2}{c}{$m=3$} \\ \cmidrule{2-8}
the slope parameter $\beta$   & \multicolumn{7}{c}{$\beta=3$}  \\
\bottomrule[0.9pt]
\end{tabular}
}}
\label{tab:train_setup}
\end{table*}


\begin{table*}[]
\centering
\caption{Test accuracy ($\%$) of prevailing methods using the ResNet-32 on imbalanced noisy CIFAR-10 and CIFAR-100. The best and the second-best performance are highlighted with \textbf{bold} and \underline{underline}, respectively. }
\small
\label{tab:imbalanced}
\scalebox{0.95}{
\setlength{\tabcolsep}{3mm}{
\begin{tabular}{c|l|ccccc|ccccc}
\toprule[0.85pt]
 \multicolumn{2}{c|}{Imbalanced ratio $\xi$}                 & \multicolumn{5}{c|}{10}      & \multicolumn{5}{c}{100} \\ \midrule
\multicolumn{2}{c|}{Noise ratio $\rho$}     & {10\%}  & {20\%} & {30\%} & {40\%} & {50\%} & {10\%}  & {20\%} & {30\%} & {40\%} & {50\%} \\ \midrule
\multirow{6}{*}{\rotatebox{90}{CIFAR-10}}   & CE    & 80.41 & 75.61 & 71.94 & 70.13 & 63.25  & 64.41 & 62.17 & 52.94 & 48.11 & 38.71\\
& Co-teaching \cite{han2018co}                                      & 80.30 & 78.54 & 68.71 & 57.10 & 46.77  & 55.58 & 50.29 & 38.01 & 30.75 & 22.85 \\
& RoLT \cite{wei2021robust}                                            & 85.68 & 85.43 & 83.50 & 80.92 & 78.96  & 73.02 & 71.20 & 66.53 & 57.86 & 48.98\\
& Sel-CL+ \cite{li2022selective}                                         & 86.47 & 85.11 & 84.41 & 80.35 & 77.27  & 72.31 & 71.02 & 65.70 & 61.37 & 56.21\\
& CurveNet \cite{jiang2022delving}  & 87.12  & 85.02   & 84.39  & 82.99 & 78.57 & 76.55 & 74.19 & 71.90 & 67.20 & 64.83 \\ 
& RCAL \cite{zhang2023noisy}    & \textbf{88.09} & \textbf{86.46} & \underline{84.58} &\underline{83.43} & \underline{80.80} & \textbf{78.60} & \underline{75.81} & \underline{72.76} & \underline{69.78} & \underline{65.05} \\ \cmidrule{2-12}
& \textbf{Ours}     & \underline{87.32} & \underline{86.31} & \textbf{84.91} & \textbf{84.72} & \textbf{81.04}  & \underline{78.39} & \textbf{76.21} & \textbf{74.19}  & \textbf{70.99} & \textbf{67.28} \\ \midrule
\multirow{6}{*}{\rotatebox{90}{CIFAR-100}}  & CE    & 48.54 & 43.27 & 37.43 & 32.94 & 26.24  & 31.81 & 26.21 & 21.79 & 17.91 & 14.23\\
& Co-teaching \cite{han2018co}                              & 45.61 & 41.33 & 36.14 & 32.08 & 25.33  & 30.55 & 25.67 & 22.01 & 16.20 & 13.45\\
& RoLT \cite{wei2021robust}                                     & 54.11 & 51.00 & 47.42 & 44.63 & 38.64  & 35.21 & 30.97 & 27.60 & 24.73 & 20.14\\
& Sel-CL+ \cite{li2022selective}                                    & 55.68 & 53.52 & 50.92 & 47.57 & \underline{44.86} & 37.45 & 36.79 & 35.09 & 31.96 & 28.59\\
& CurveNet \cite{jiang2022delving}      & 55.96 & 54.60 & 51.28 & 48.10 & 44.61 & 40.91 & 39.71 & 35.90 & 32.09 & 29.71 \\
& RCAL \cite{zhang2023noisy}   & \textbf{57.50} & \underline{54.85} & \underline{51.66} & \textbf{48.91} & 44.36 & \underline{41.68} & \underline{39.85} & \underline{36.57} & \underline{33.36}  & \underline{30.26}\\ \cmidrule{2-12}
& \textbf{Ours}                             & \underline{56.10} & \textbf{55.93} & \textbf{53.07} & \underline{48.29} & \textbf{45.17} & \textbf{44.00}  & \textbf{41.90} & \textbf{40.44} & \textbf{37.28} & \textbf{34.80}\\ 
\bottomrule[0.85pt]  
\end{tabular}}
}
\vspace{-1mm}
\end{table*}

\section{More Experimental Results}

\subsection{Results on Imbalanced Noisy Labels}
Recently, some literature \cite{wei2021robust,zhang2023noisy,jiang2022delving} studied a more challenging noise learning task, i.e., learning with noisy labels on imbalanced datasets, which is a more realistic scenario. Our method can also tackle this task thanks to the class-aware data sampling strategy in ITEM. Following the imbalanced noise setting and experimental setup of RCAL \cite{zhang2023noisy}, we verify the performance of $\ddagger$ITEM on comprehensive conditions. 

\textbf{Dataset construction.} Let $T_{ij}(\bm{x})$ denote a probability that the true label $i$ of the instance $\bm{x}$ is corrupted to the noisy label. Therefore, given a noise ratio $rho$, the noise transition matrix satisfies that $T_{ij}(\bm{x}) = \rho$ if $i=j$, otherwise $T_{ij}(\bm{x}) = \frac{1}{C-1}\rho$, where $C$ denotes the number of classes. We first construct an imbalanced dataset on CIFAR with different imbalanced ratios $\xi = \{10, 100\}$ and then corrupt the labels with varying noisy ratio $\rho = \{10\%,...,50\%\}$.

We compare our proposal ITEM with various methods in LNL and imbalanced LNL, including Co-teaching \cite{han2018co}, RoLT \cite{wei2021robust}, Sel-CL+ \cite{li2022selective}, and RCAL \cite{zhang2023noisy}. Comparison results are shown in Table \ref{tab:imbalanced}. Weighted sampling, which mitigates data bias by adjusting selection probabilities and increasing the visibility of underrepresented classes during training, helps ITEM tackle long-tailed classification tasks.
On both imbalanced noisy datasets, ITEM exhibits greater generalization under a high noise ratio. For example, ITEM outperforms RCAL by 1.29\% and 0.24\% under 40\% and 50\% noise ratio, respectively. On CIFAR-100 with an imbalanced ratio $\xi=100$, ITEM consistently achieves the best generalization performance over other comparison methods. Under an extreme noise rate $\rho=50\%$, ITEM achieves more than 4.5\% improvement of top-1 test accuracy. Therefore, we believe that ITEM can also fulfill strong robustness even under imbalanced real-world noisy conditions.

\subsection{Visualization of Class-Level Selection}
\begin{figure}[t]
    \centering
    \includegraphics[width=0.5\linewidth]{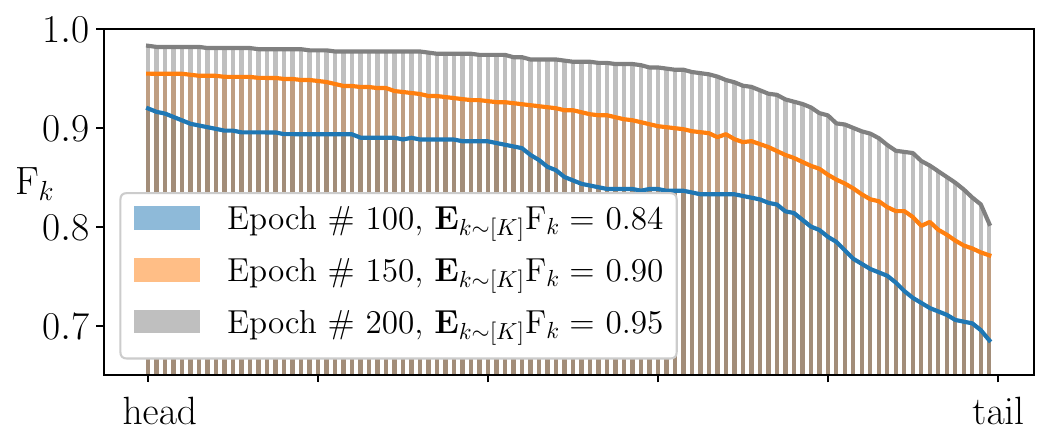}
    \caption{\textbf{Data debias} on selection. We plot the class-level selection performance of ITEM on CIFAR-100 with 60\% symmetric label noise.} \vspace{-2mm}
    \label{fig:debias_Fscore}
\end{figure}

As shown in Figure \ref{fig:debias_Fscore}, we further visualize the class-level selection performance of our proposal. We can see that the total selection performance increases as the training progresses. Further, in the tailed classes where the value $F_k$ in other selection frameworks would decrease, ITEM gradually improves the F-score in all classes instead of only in head classes, demonstrating that ITEM effectively mitigates the data bias.

\begin{table*}[t]
\centering
\caption{Comparison of total training time in hours on CIFAR10N with \textit{Worst} noise labels with varying experts number.}
\small
\resizebox{0.85\textwidth}{!}{
\setlength{\tabcolsep}{2mm}{
\begin{tabular}{l|ccc|cccc}
\toprule[0.9pt]
\multirow{3}{*}{Metric} & \multicolumn{1}{c}{Single Network}  & \multicolumn{2}{c|}{Double-branch Network}   & \multicolumn{4}{c}{\textbf{Ours: MoE}}\\ \cmidrule{2-8}
                            & CE       & Co-teaching& DivideMix     & ITEM    & ITEM    & ITEM*   & ITEM* \\ 
    &   &{\scriptsize \cite{han2018co}}  & {\scriptsize \cite{li2020dividemix}}   &  {\scriptsize ($m=3$)}&  {\scriptsize ($m=5$)}&  {\scriptsize ($m=3$)}&  {\scriptsize ($m=5$)} \\ \midrule
Test accuracy (\%) $\uparrow$    & 77.69  & 83.26    &  92.56      & 91.15       & 91.40       & 93.14       & 93.32 \\ 
Times (hours) $\downarrow$ & 1.6       & 4.2        &  4.4          & 1.6           & 1.7          & 1.8           & 1.8 \\
\bottomrule[0.9pt]  
\end{tabular}}
}
\label{tab:computation}
\end{table*}

\begin{table*}[t]
\centering
\small
\caption{Test accuracy (\%) on two fine-grained benchmarks with two noisy settings. The best and the second best performances are highlighted with \textbf{Bold} and \underline{underline}, respectively.}
\resizebox{0.85\textwidth}{!}{
\setlength{\tabcolsep}{4mm}{
\begin{tabular}{l|cccc|c}
\toprule[0.9pt]
\multirow{2}{*}{Methods}     & \multicolumn{2}{c}{Stanford Dogs}    & \multicolumn{2}{c|}{CUB-200-2011}  & \multirow{2}{*}{\emph{Avg.}} \\
& {\small Sym 40\%}     & {\small Asy 30\%}    
& {\small Sym 40\%}     & {\small Asy 30\%}  \\ \midrule
Cross-Entropy                     & 51.42      & 58.08     & 64.01     & 56.02     & 57.38 \\
Confidence Penalty \cite{pereyra2017regularizing}         & 68.69      & 64.50     & 52.40     & 54.33     & 59.98 \\
Label Smooth \cite{lukasik2020does}           & 70.22      & \underline{64.99}     & 54.39     & 56.80     & 61.60 \\
Co-teaching \cite{han2018co}           & 49.15      & 50.50     & 46.57     & 50.60     & 49.21 \\
JoCoR  \cite{wei2020combating}                & 49.62      & 53.59     & 52.64     & 51.70     & 51.89 \\
SNSCL \cite{wei2023fine}            & \textbf{75.27}      & 64.49     & \underline{68.83}     & \underline{61.48}     & \underline{67.52} \\ \midrule
Ours        & \underline{74.92}      & \textbf{68.19}     & \textbf{70.04}     & \textbf{66.58}     & \textbf{69.93} \\
\bottomrule[0.9pt]
\end{tabular}
}}
\label{tab:finegrained}
\end{table*}

\subsection{Training Time Analysis}
In Table \ref{tab:computation}, we compare the training times of ITEM and (semi) ITEM* with three typical methods, using a single Nvidia 4090 GPU. We can observe that the training time of these methods based on double-brach networks (such as Co-teaching and DivideMix) is always twice as slow as training on a single network. However, in our proposed MoE structure, increasing the number of experts number will not significantly increase the computation cost. Besides, integrating ITEM with a semi-supervised learning framework can bring considerable performance improvement with fewer extra costs.

\subsection{Results on Fine-Grained Noisy Labels}
Recently, one literature \cite{wei2023fine} pointed out that noisy label generation is strongly associated with fine-grained datasets. Considering the property of wide existence and strong meaningfulness of noisy fine-grained classification, following settings in SNSCL \cite{wei2023fine}, we conduct experiments on four noisy fine-grained datasets, including Stanford Dogs \cite{FeiFei_FGVC2011} and CUB-200-2011 \cite{Caltech399}. Given the selection criterion GMM-based small loss selection, we fine-tuned a (pre-trained) ResNet-18 with three experts. We synthetically construct two noise types, symmetric and asymmetric noise labels. 

The results are shown in Table \ref{tab:finegrained}. Our method achieves significant generalization performance even when tackling fine-grained noisy classification and consistently outperforms other methods by a large margin. For example, under Stanford Dogs with 30\% asymmetric noise, ITEM improves the best method (label smooth) by more than 4\%.
We consider the MixUp operation to be the primary contribution to the advanced performance on fine-grained datasets for two reasons, 1) it encourages the model to learn from structured data instead of the unstructured noise \cite{arazo2019unsupervised}, and 2) it encourages the model to learn a more generalized decision boundary, which largely benefits the fine-grained classification task.

%% file: main.bbl
\begin{thebibliography}{58}
\providecommand{\natexlab}[1]{#1}
\providecommand{\url}[1]{\texttt{#1}}
\expandafter\ifx\csname urlstyle\endcsname\relax
  \providecommand{\doi}[1]{doi: #1}\else
  \providecommand{\doi}{doi: \begingroup \urlstyle{rm}\Url}\fi

\bibitem[Arazo et~al.(2019)Arazo, Ortego, Albert, O’Connor, and McGuinness]{arazo2019unsupervised}
Eric Arazo, Diego Ortego, Paul Albert, Noel O’Connor, and Kevin McGuinness.
\newblock Unsupervised label noise modeling and loss correction.
\newblock In \emph{ICML}, 2019.

\bibitem[Arpit et~al.(2017)Arpit, Jastrzkebski, Ballas, Krueger, Bengio, Kanwal, Maharaj, Fischer, Courville, Bengio, et~al.]{arpit2017closer}
Devansh Arpit, Stanislaw Jastrzkebski, Nicolas Ballas, David Krueger, Emmanuel Bengio, Maxinder~S Kanwal, Tegan Maharaj, Asja Fischer, Aaron Courville, Yoshua Bengio, et~al.
\newblock A closer look at memorization in deep networks.
\newblock In \emph{ICML}, 2017.

\bibitem[Bai and Liu(2021)]{bai2021me}
Yingbin Bai and Tongliang Liu.
\newblock Me-momentum: Extracting hard confident examples from noisily labeled data.
\newblock In \emph{ICCV}, 2021.

\bibitem[Bai et~al.(2021)Bai, Yang, Han, Yang, Li, Mao, Niu, and Liu]{bai2021understanding}
Yingbin Bai, Erkun Yang, Bo Han, Yanhua Yang, Jiatong Li, Yinian Mao, Gang Niu, and Tongliang Liu.
\newblock Understanding and improving early stopping for learning with noisy labels.
\newblock In \emph{NeurIPS}, 2021.

\bibitem[Bengio et~al.(2009)Bengio, Louradour, Collobert, and Weston]{bengio2009curriculum}
Yoshua Bengio, J{\'e}r{\^o}me Louradour, Ronan Collobert, and Jason Weston.
\newblock Curriculum learning.
\newblock In \emph{ICML}, 2009.

\bibitem[Blum et~al.(2003)Blum, Kalai, and Wasserman]{blum2003noise}
Avrim Blum, Adam Kalai, and Hal Wasserman.
\newblock Noise-tolerant learning, the parity problem, and the statistical query model.
\newblock \emph{JACM}, 2003.

\bibitem[Bossard et~al.(2014)Bossard, Guillaumin, and Gool]{bossard2014food}
Lukas Bossard, Matthieu Guillaumin, and Luc~Van Gool.
\newblock Food-101--mining discriminative components with random forests.
\newblock In \emph{ECCV}, 2014.

\bibitem[Cheng et~al.(2021)Cheng, Zhu, Li, Gong, Sun, and Liu]{cheng2020learning}
Hao Cheng, Zhaowei Zhu, Xingyu Li, Yifei Gong, Xing Sun, and Yang Liu.
\newblock Learning with instance-dependent label noise: A sample sieve approach.
\newblock In \emph{ICLR}, 2021.

\bibitem[Dong et~al.(2020)Dong, Yu, Cao, Shi, and Ma]{dong2020survey}
Xibin Dong, Zhiwen Yu, Wenming Cao, Yifan Shi, and Qianli Ma.
\newblock A survey on ensemble learning.
\newblock \emph{FCS}, 2020.

\bibitem[Han et~al.(2018{\natexlab{a}})Han, Yao, Niu, Zhou, Tsang, Zhang, and Sugiyama]{han2018masking}
Bo Han, Jiangchao Yao, Gang Niu, Mingyuan Zhou, Ivor Tsang, Ya Zhang, and Masashi Sugiyama.
\newblock Masking: A new perspective of noisy supervision.
\newblock In \emph{NeurIPS}, 2018{\natexlab{a}}.

\bibitem[Han et~al.(2018{\natexlab{b}})Han, Yao, Yu, Niu, Xu, Hu, Tsang, and Sugiyama]{han2018co}
Bo Han, Quanming Yao, Xingrui Yu, Gang Niu, Miao Xu, Weihua Hu, Ivor Tsang, and Masashi Sugiyama.
\newblock Co-teaching: Robust training of deep neural networks with extremely noisy labels.
\newblock In \emph{NeurIPS}, 2018{\natexlab{b}}.

\bibitem[Jiang et~al.(2018)Jiang, Zhou, Leung, Li, and Fei-Fei]{jiang2018mentornet}
Lu Jiang, Zhengyuan Zhou, Thomas Leung, Li-Jia Li, and Li Fei-Fei.
\newblock Mentornet: Learning data-driven curriculum for very deep neural networks on corrupted labels.
\newblock In \emph{ICML}, 2018.

\bibitem[Jiang et~al.(2022)Jiang, Li, Wang, Huang, Zhang, and Xu]{jiang2022delving}
Shenwang Jiang, Jianan Li, Ying Wang, Bo Huang, Zhang Zhang, and Tingfa Xu.
\newblock Delving into sample loss curve to embrace noisy and imbalanced data.
\newblock In \emph{AAAI}, 2022.

\bibitem[Jindal et~al.(2016)Jindal, Nokleby, and Chen]{jindal2016learning}
Ishan Jindal, Matthew Nokleby, and Xuewen Chen.
\newblock Learning deep networks from noisy labels with dropout regularization.
\newblock In \emph{ICDM}, 2016.

\bibitem[Khosla et~al.(2011)Khosla, Jayadevaprakash, Yao, and Fei-Fei]{FeiFei_FGVC2011}
Aditya Khosla, Nityananda Jayadevaprakash, Bangpeng Yao, and Li Fei-Fei.
\newblock Novel dataset for fine-grained image categorization.
\newblock In \emph{CVPR Workshop}, 2011.

\bibitem[Kotsiantis et~al.(2006)Kotsiantis, Zaharakis, and Pintelas]{kotsiantis2006machine}
Sotiris~B Kotsiantis, Ioannis~D Zaharakis, and Panayiotis~E Pintelas.
\newblock Machine learning: a review of classification and combining techniques.
\newblock \emph{Artificial Intelligence Review}, 2006.

\bibitem[Krizhevsky et~al.(2009)Krizhevsky, Hinton, et~al.]{krizhevsky2009learning}
Alex Krizhevsky, Geoffrey Hinton, et~al.
\newblock Learning multiple layers of features from tiny images.
\newblock 2009.

\bibitem[Krizhevsky et~al.(2012)Krizhevsky, Sutskever, and Hinton]{krizhevsky2012imagenet}
Alex Krizhevsky, Ilya Sutskever, and Geoffrey~E Hinton.
\newblock Imagenet classification with deep convolutional neural networks.
\newblock In \emph{NeurIPS}, 2012.

\bibitem[Lee et~al.(2013)]{lee2013pseudo}
Dong-Hyun Lee et~al.
\newblock Pseudo-label: The simple and efficient semi-supervised learning method for deep neural networks.
\newblock In \emph{ICML Workshop}, 2013.

\bibitem[Lee et~al.(2018)Lee, He, Zhang, and Yang]{lee2018cleannet}
Kuang-Huei Lee, Xiaodong He, Lei Zhang, and Linjun Yang.
\newblock Cleannet: Transfer learning for scalable image classifier training with label noise.
\newblock In \emph{CVPR}, 2018.

\bibitem[Li et~al.(2020)Li, Socher, and Hoi]{li2020dividemix}
Junnan Li, Richard Socher, and Steven~C.H. Hoi.
\newblock Dividemix: Learning with noisy labels as semi-supervised learning.
\newblock In \emph{ICLR}, 2020.

\bibitem[Li et~al.(2022)Li, Xia, Ge, and Liu]{li2022selective}
Shikun Li, Xiaobo Xia, Shiming Ge, and Tongliang Liu.
\newblock Selective-supervised contrastive learning with noisy labels.
\newblock In \emph{CVPR}, 2022.

\bibitem[Li et~al.(2017)Li, Wang, Li, Agustsson, and Van~Gool]{li2017webvision}
Wen Li, Limin Wang, Wei Li, Eirikur Agustsson, and Luc Van~Gool.
\newblock Webvision database: Visual learning and understanding from web data.
\newblock \emph{arXiv:1708.02862}, 2017.

\bibitem[Li et~al.(2023)Li, Han, Shan, and Chen]{li2023disc}
Yifan Li, Hu Han, Shiguang Shan, and Xilin Chen.
\newblock Disc: Learning from noisy labels via dynamic instance-specific selection and correction.
\newblock In \emph{CVPR}, 2023.

\bibitem[Li et~al.(2024)Li, Chen, Yu, Chen, and Shen]{li2024sure}
Yuting Li, Yingyi Chen, Xuanlong Yu, Dexiong Chen, and Xi Shen.
\newblock Sure: Survey recipes for building reliable and robust deep networks.
\newblock In \emph{CVPR}, 2024.

\bibitem[Liu et~al.(2020)Liu, Niles-Weed, Razavian, and Fernandez-Granda]{liu2020early}
Sheng Liu, Jonathan Niles-Weed, Narges Razavian, and Carlos Fernandez-Granda.
\newblock Early-learning regularization prevents memorization of noisy labels.
\newblock In \emph{NeurIPS}, 2020.

\bibitem[Liu and Guo(2020)]{liu2020peer}
Yang Liu and Hongyi Guo.
\newblock Peer loss functions: Learning from noisy labels without knowing noise rates.
\newblock In \emph{ICML}, 2020.

\bibitem[Lukasik et~al.(2020)Lukasik, Bhojanapalli, Menon, and Kumar]{lukasik2020does}
Michal Lukasik, Srinadh Bhojanapalli, Aditya Menon, and Sanjiv Kumar.
\newblock Does label smoothing mitigate label noise?
\newblock In \emph{ICML}, 2020.

\bibitem[Masoudnia and Ebrahimpour(2014)]{masoudnia2014mixture}
Saeed Masoudnia and Reza Ebrahimpour.
\newblock Mixture of experts: a literature survey.
\newblock \emph{Artificial Intelligence Review}, 2014.

\bibitem[Pereyra et~al.(2017)Pereyra, Tucker, Chorowski, Kaiser, and Hinton]{pereyra2017regularizing}
Gabriel Pereyra, George Tucker, Jan Chorowski, {\L}ukasz Kaiser, and Geoffrey Hinton.
\newblock Regularizing neural networks by penalizing confident output distributions.
\newblock \emph{arXiv:1701.06548}, 2017.

\bibitem[Rokach(2010)]{rokach2010ensemble}
Lior Rokach.
\newblock Ensemble-based classifiers.
\newblock \emph{Artificial intelligence review}, 2010.

\bibitem[Sohn et~al.(2020)Sohn, Berthelot, Carlini, Zhang, Zhang, Raffel, Cubuk, Kurakin, and Li]{sohn2020fixmatch}
Kihyuk Sohn, David Berthelot, Nicholas Carlini, Zizhao Zhang, Han Zhang, Colin~A Raffel, Ekin~Dogus Cubuk, Alexey Kurakin, and Chun-Liang Li.
\newblock Fixmatch: Simplifying semi-supervised learning with consistency and confidence.
\newblock In \emph{NeurIPS}, 2020.

\bibitem[Szegedy et~al.(2017)Szegedy, Ioffe, Vanhoucke, and Alemi]{szegedy2017inception}
Christian Szegedy, Sergey Ioffe, Vincent Vanhoucke, and Alexander Alemi.
\newblock Inception-v4, inception-resnet and the impact of residual connections on learning.
\newblock In \emph{AAAI}, 2017.

\bibitem[Tanaka et~al.(2018)Tanaka, Ikami, Yamasaki, and Aizawa]{tanaka2018joint}
Daiki Tanaka, Daiki Ikami, Toshihiko Yamasaki, and Kiyoharu Aizawa.
\newblock Joint optimization framework for learning with noisy labels.
\newblock In \emph{CVPR}, 2018.

\bibitem[Van~der Maaten and Hinton(2008)]{van2008visualizing}
Laurens Van~der Maaten and Geoffrey Hinton.
\newblock Visualizing data using t-sne.
\newblock \emph{JMLR}, 2008.

\bibitem[Wang et~al.(2022)Wang, Xiao, Dong, Feng, and Zhao]{wang2022promix}
Haobo Wang, Ruixuan Xiao, Yiwen Dong, Lei Feng, and Junbo Zhao.
\newblock Promix: Combating label noise via maximizing clean sample utility.
\newblock In \emph{IJCAI}, 2022.

\bibitem[Wei et~al.(2020)Wei, Feng, Chen, and An]{wei2020combating}
Hongxin Wei, Lei Feng, Xiangyu Chen, and Bo An.
\newblock Combating noisy labels by agreement: A joint training method with co-regularization.
\newblock In \emph{CVPR}, 2020.

\bibitem[Wei et~al.(2021{\natexlab{a}})Wei, Zhu, Cheng, Liu, Niu, and Liu]{wei2021learning}
Jiaheng Wei, Zhaowei Zhu, Hao Cheng, Tongliang Liu, Gang Niu, and Yang Liu.
\newblock Learning with noisy labels revisited: A study using real-world human annotations.
\newblock \emph{arXiv:2110.12088}, 2021{\natexlab{a}}.

\bibitem[Wei et~al.(2022)Wei, Sun, Lu, and Yin]{wei2022self}
Qi Wei, Haoliang Sun, Xiankai Lu, and Yilong Yin.
\newblock Self-filtering: A noise-aware sample selection for label noise with confidence penalization.
\newblock In \emph{ECCV}, 2022.

\bibitem[Wei et~al.(2023)Wei, Feng, Sun, Wang, Guo, and Yin]{wei2023fine}
Qi Wei, Lei Feng, Haoliang Sun, Ren Wang, Chenhui Guo, and Yilong Yin.
\newblock Fine-grained classification with noisy labels.
\newblock In \emph{CVPR}, 2023.

\bibitem[Wei et~al.(2021{\natexlab{b}})Wei, Shi, Tu, and Li]{wei2021robust}
Tong Wei, Jiang-Xin Shi, Wei-Wei Tu, and Yu-Feng Li.
\newblock Robust long-tailed learning under label noise.
\newblock \emph{arXiv:2108.11569}, 2021{\natexlab{b}}.

\bibitem[Welinder et~al.(2010)Welinder, Branson, Mita, Wah, Schroff, Belongie, and Perona]{Caltech399}
Peter Welinder, Steve Branson, Takeshi Mita, Catherine Wah, Florian Schroff, Serge Belongie, and Pietro Perona.
\newblock Caltech-ucsd birds 200.
\newblock Technical report, 2010.

\bibitem[Xia et~al.(2020)Xia, Liu, Han, Gong, Wang, Ge, and Chang]{xia2020robust}
Xiaobo Xia, Tongliang Liu, Bo Han, Chen Gong, Nannan Wang, Zongyuan Ge, and Yi Chang.
\newblock Robust early-learning: Hindering the memorization of noisy labels.
\newblock In \emph{ICLR}, 2020.

\bibitem[Xia et~al.(2022)Xia, Liu, Han, Gong, Yu, Niu, and Sugiyama]{xia2021sample}
Xiaobo Xia, Tongliang Liu, Bo Han, Mingming Gong, Jun Yu, Gang Niu, and Masashi Sugiyama.
\newblock Sample selection with uncertainty of losses for learning with noisy labels.
\newblock In \emph{ICLR}, 2022.

\bibitem[Xia et~al.(2023)Xia, Han, Zhan, Yu, Gong, Gong, and Liu]{xia2023combating}
Xiaobo Xia, Bo Han, Yibing Zhan, Jun Yu, Mingming Gong, Chen Gong, and Tongliang Liu.
\newblock Combating noisy labels with sample selection by mining high-discrepancy examples.
\newblock In \emph{ICCV}, 2023.

\bibitem[Xiao et~al.(2015)Xiao, Xia, Yang, Huang, and Wang]{xiao2015learning}
Tong Xiao, Tian Xia, Yi Yang, Chang Huang, and Xiaogang Wang.
\newblock Learning from massive noisy labeled data for image classification.
\newblock In \emph{CVPR}, 2015.

\bibitem[Xu et~al.(2019)Xu, Cao, Kong, and Wang]{xu2019l_dmi}
Yilun Xu, Peng Cao, Yuqing Kong, and Yizhou Wang.
\newblock L\_dmi: A novel information-theoretic loss function for training deep nets robust to label noise.
\newblock In \emph{NeurIPS}, 2019.

\bibitem[Yan et~al.(2014)Yan, Rosales, Fung, Subramanian, and Dy]{yan2014learning}
Yan Yan, R{\'o}mer Rosales, Glenn Fung, Ramanathan Subramanian, and Jennifer Dy.
\newblock Learning from multiple annotators with varying expertise.
\newblock \emph{MLJ}, 2014.

\bibitem[Yao et~al.(2018)Yao, Wang, Tsang, Zhang, Sun, Zhang, and Zhang]{yao2018deep}
Jiangchao Yao, Jiajie Wang, Ivor~W Tsang, Ya Zhang, Jun Sun, Chengqi Zhang, and Rui Zhang.
\newblock Deep learning from noisy image labels with quality embedding.
\newblock \emph{IEEE TIP}, 2018.

\bibitem[Yu et~al.(2019)Yu, Han, Yao, Niu, Tsang, and Sugiyama]{yu2019does}
Xingrui Yu, Bo Han, Jiangchao Yao, Gang Niu, Ivor Tsang, and Masashi Sugiyama.
\newblock How does disagreement help generalization against label corruption?
\newblock In \emph{ICML}, 2019.

\bibitem[Yuan et~al.(2023)Yuan, Feng, and Liu]{yuan2023late}
Suqin Yuan, Lei Feng, and Tongliang Liu.
\newblock Late stopping: Avoiding confidently learning from mislabeled examples.
\newblock In \emph{ICCV}, 2023.

\bibitem[Zhang et~al.(2017{\natexlab{a}})Zhang, Bengio, Hardt, Recht, and Vinyals]{zhang2017understanding}
Chiyuan Zhang, Samy Bengio, Moritz Hardt, Benjamin Recht, and Oriol Vinyals.
\newblock Understanding deep learning requires rethinking generalization.
\newblock In \emph{ICLR}, 2017{\natexlab{a}}.

\bibitem[Zhang et~al.(2017{\natexlab{b}})Zhang, Cisse, Dauphin, and Lopez-Paz]{zhang2017mixup}
Hongyi Zhang, Moustapha Cisse, Yann~N Dauphin, and David Lopez-Paz.
\newblock mixup: Beyond empirical risk minimization.
\newblock \emph{arXiv:1710.09412}, 2017{\natexlab{b}}.

\bibitem[Zhang et~al.(2023)Zhang, Zhao, Yao, Yuan, and Huang]{zhang2023noisy}
Manyi Zhang, Xuyang Zhao, Jun Yao, Chun Yuan, and Weiran Huang.
\newblock When noisy labels meet long tail dilemmas: A representation calibration method.
\newblock In \emph{ICCV}, 2023.

\bibitem[Zhou et~al.(2020{\natexlab{a}})Zhou, Cui, Wei, and Chen]{zhou2020bbn}
Boyan Zhou, Quan Cui, Xiu-Shen Wei, and Zhao-Min Chen.
\newblock Bbn: Bilateral-branch network with cumulative learning for long-tailed visual recognition.
\newblock In \emph{CVPR}, 2020{\natexlab{a}}.

\bibitem[Zhou et~al.(2020{\natexlab{b}})Zhou, Wang, and Bilmes]{zhou2020robust}
Tianyi Zhou, Shengjie Wang, and Jeff Bilmes.
\newblock Robust curriculum learning: From clean label detection to noisy label self-correction.
\newblock In \emph{ICLR}, 2020{\natexlab{b}}.

\bibitem[Zhu et~al.(2021)Zhu, Liu, and Liu]{zhu2021second}
Zhaowei Zhu, Tongliang Liu, and Yang Liu.
\newblock A second-order approach to learning with instance-dependent label noise.
\newblock In \emph{CVPR}, 2021.

\bibitem[Zong et~al.(2024)Zong, Wang, Xie, and Huang]{zong2024dirichlet}
Chen-Chen Zong, Ye-Wen Wang, Ming-Kun Xie, and Sheng-Jun Huang.
\newblock Dirichlet-based prediction calibration for learning with noisy labels.
\newblock In \emph{AAAI}, 2024.

\end{thebibliography}
